\documentclass{article} 
\usepackage{nips14submit_e,times}
\usepackage{hyperref}
\usepackage{url}
\usepackage{visionlab,xspace,tabularx}
\usepackage{graphicx}
\usepackage{epstopdf}
\usepackage{caption} 
\usepackage{subcaption}
\usepackage{amsmath,amssymb} 

\title{D-MFVI: Distributed Mean Field Variational Inference using Bregman ADMM}

\author{
Behnam Babagholami-Mohamadabadi\\
Department of Computer Science\\
Rutgers, The State University of New Jersey\\
Piscataway, NJ \\
\texttt{bb510@cs.rutgers.edu} \\
\And
Sejong Yoon \\
Department of Computer Science\\
Rutgers, The State University of New Jersey\\
Piscataway, NJ \\
\texttt{sjyoon@cs.rutgers.edu} \\
\AND
Vladimir Pavlovic \\
Department of Computer Science\\
Rutgers, The State University of New Jersey\\
Piscataway, NJ \\
\texttt{vladimir@cs.rutgers.edu} \\
}

\nipsfinalcopy 

\begin{document}

\maketitle

\begin{abstract}
Bayesian models provide a framework for probabilistic modelling of complex datasets. However, many of such models are computationally demanding especially in the presence of large datasets. On the other hand, in sensor network applications, statistical (Bayesian) parameter estimation usually needs distributed algorithms, in which both data and computation are distributed across the nodes of the network.
In this paper we propose a general framework for distributed Bayesian learning using Bregman Alternating Direction Method of Multipliers (B-ADMM). We demonstrate the utility of our framework, with Mean Field Variational Bayes (MFVB) as the primitive for distributed Matrix Factorization (MF) and distributed affine structure from motion (SfM). 
\end{abstract}

\section{Introduction}
Traditional setting for many machine learning algorithms is the one where the model (e.g., a classifier or a regressor, typically parametric in some sense) is constructed from a body of data by processing this body in either batch or online fashion.  The model itself is centralized and the algorithm has access to all model parameters and all data points.  However, in many application scenarios today it is not reasonable to assume access to all data points because they could be distributed over a network of sensor or processing nodes.  In those settings collecting and processing data in a centralized fashion is not always feasible because of several important challenges.  First, in applications such as networks of cameras mounted on vehicles, the networks are constrained by severe capacity and energy constraints, considerably limiting the node communications [1,5].  Second, in many distributed sensor network  applications such as health care, ecological monitoring, or smart homes, collecting all data at a single location may not be feasible because of its sheer volume as well as potential privacy concerns. Lastly, the size of the centralized data would incur an insurmountable computational burden on the algorithm, preventing real-time or anytime [24] processing often desired in large sensing systems [4].\\

Distributed sensor networks provide an application setting in which distributed optimization tasks (including machine learning) that deal with some of the aforementioned challenges are frequently addressed [1,17].  However, they are traditionally considered in a non-Bayesian (often deterministic) fashion.  Moreover, the data is often assumed to be complete (not missing) across the network and in individual nodes.  As a consequence, these approaches usually obtain parameter point estimates by minimizing a loss function based on the complete data and dividing the computation into subset-specific optimization problems. A more challenging yet critical problem is to provide full posterior distributions for those parameters. Such posteriors have the major advantage of characterizing the uncertainty in parameter learning and predictions, absent from traditional distributed optimization approaches. Another drawback of such approaches is that they traditionally rely on batch processing within individual nodes, unable to seamlessly deal with streaming data frequently present in sensing networks. Moreover, the deterministic distributed optimization approaches are not inherently able to account for various forms of the missing data, including missing at random or not-at-random [15].  However, both the sequential inference and the data completion would be naturally handled via the Bayesian analysis [3] if one could obtain full posterior parameter estimates in this distributed setting.\\

In a recent work [2] proposed a new method that estimates parametric probabilistic models with latent variables in a distributed network setting.  The performance of this model was demonstrated to be on par with the centralized model, while it could efficiently deal with the distributed missing data. Nevertheless, the approach has several drawbacks.  First, its use of the Maximum Likelihood (ML) estimation increases the risk of overfitting, which is particularly pronounced in the distributed setting where each node works with a subset of the full data. Second, the approach cannot provide the uncertainty around the estimated parameters that may be crucial in many applications, e.g., in online learning for streaming data or in assessing confidence of predictions.  In this paper we propose a Distributed Mean Field Variational Inference (D-MFVI) algorithm for Bayesian Inference in a large class of graphical models. The goal of our framework is to learn a single consensus Bayesian model by doing local Bayesian inference and in-network information sharing without the need for centralized computation and/or centralized data gathering. In particular, we demonstrate D-MFVI on the Bayesian Principle Component Analysis (BPCA) problem and then apply this model to solve two distributed Bayesian matrix factorization tasks, a matrix completion for item rating/preference modeling and the distributed structure-from-motion task in a camera network. \\

The reminder of this paper is as follows. In Section~\ref{pm}, we briefly review MFVI for graphical models and then derive distributed MFVB. In Section~\ref{BPCA}, we review Bayesian PCA and then derive D-MFVI in this setting.
 In Section~\ref{er}, we report experimental results of our model using both synthetic and real data. Finally, we discuss our approach including its limitations and possible solutions in Section~\ref{co}.
\section{Distributed Mean Field Variational Inference (D-MFVI)}\label{pm}
We first explain a general parametric Bayesian model in a centralized setting and then we derive its distributed form.
\subsection{Centralized Setting}
\begin{figure}[t]
\begin{center}
\includegraphics[width=0.4\linewidth]{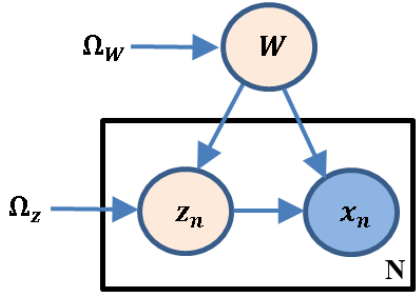}
\end{center}
\caption{A graphical representation of the model of Eq. (\ref{2}) (blue-shaded circle denotes observation).}
\label{fig:1}
\end{figure}

Consider a data set $X$ of observed D-dimensional vectors $X = \{x_i \in \mathbb{R}^D\}_{i=1}^{N}$ with the corresponding local latent variables $Z=\{z_i \in \mathbb{R}^{M}\}_{i=1}^{N}$, a global latent variable $W \in \mathbb{R}^{p}$ and a set of fixed parameters $\Omega=[\Omega_z, \Omega_w]$. The main assumption of our class of models is the factorization of the joint distribution of observations, global and local variables into a global term and a product of local terms (the graphical representation of this class of models is shown in Fig \ref{fig:1}):
\begin{equation}\label{2}
P(X,Z,W|\Omega) = P(W|\Omega_w) \prod_{i=1}^{N}P(x_n|z_n,W)P(z_n|\Omega_z)
\end{equation}  
Given the observation, the goal is to compute the approximation of the posterior distribution of the latent variables: $P(W,Z|X,\Omega)$. For ease of computation, we consider exponential family assumption of the conditional distribution of a latent variable given the observation and the other latent variables:
 \begin{equation}\label{4}
 P(W|X,Z,\Omega_w) = h(W)exp\big\{ \eta_w(X,Z,\Omega_w)^{\top}\mathcal{T}(W) - \mathcal{A}_w\big(\eta_w(X,Z,\Omega_w)  \big) \big\}
 \end{equation}
 \begin{equation}\label{5}
 P(Z|X,W,\Omega_z) =\prod_{n=1}^{N} h(z_n)exp\big\{ \eta_{z_{n}}(x_n,W, \Omega_z)^{\top}\mathcal{T}(z_n) - \mathcal{A}_{z}\big(\eta_{z_{n}}(X,W, \Omega_z)  \big) \big\}
 \end{equation}
 where $h(.)$ denotes the \textit{base measure}, $\mathcal{A}(.)$ denotes the \textit{log partition function}, and $\eta(.)$ and $\mathcal{T}(.)$ denote the \textit{natural parameter} and the \textit{sufficient statistics} respectively. The assumed class of models contains many well known statistical models such as Bayesian PCA and Bayesian Mixture of PCA [8], Latent Dirichlet Allocation [6], Bayesian Gaussian Mixture model [7], Hidden Markov model [9,10], etc.
 
 In many probabilistic models, due to the intractability of computing the exact posterior distribution of the latent variables given the observations, we have to use approximate inference algorithms among which we use MFVI, which roots our strategy for distributed inference.  
 Using conjugate priors for $W$ and $Z$ corresponding to exponential family of the conditional distributions, MFVI always returns a distribution in the same exponential family of the prior distributions. 
 \subsection{Mean Field Variational Inference (MFVI)}
 The goal of the Variational Inference is to approximate the true posterior distribution over the latent variables with a simpler distribution indexed by a set of free parameters that is closest in KL divergence to the true posterior distribution [16]. MFVI is a subclass of VI that uses a family where all latent variables are independent of each other. More precisely, MFVI considers the following family of distributions as the approximate posterior distribution.
 \begin{equation}
 Q(Z,W) = \prod_{n=1}^{N}Q(z_n;\lambda_{z_n})Q(W;\lambda_W)
 \end{equation} 
 where the form of $Q(z_n;\lambda_{z_n})$ and $Q(W;\lambda_{W})$ are set to be in the same exponential family as the conditional distributions $P(W|X,Z,\Omega_w)$ (Eq. \ref{4}) and $P(Z|X,W,\Omega_z)$ (Eq. \ref{5}) and $\lambda_Z = \{\lambda_{z_n} \}_{n=1}^{N}$ and $\lambda_{W}$ denote the variational parameters that are determined by maximizing the following variational objective function (that is equivalent to minimizing the $KL(Q(Z,W)||P(Z,W|X,\Omega_z,\Omega_w)$) [23].
 \begin{align}
 \mathcal{L}(\lambda_Z,\lambda_W) &= \mathbb{E}_Q\big[ \log P(X,Z,W|\Omega_z,\Omega_w) \big] - \mathbb{E}_Q[\log Q] \nonumber \\
 &= \sum_{n=1}^{N} \mathbb{E}_{Q(z_n,W)}\big[ \log P(x_n|z_n,W) \big]  + \sum_{n=1}^{N} \mathbb{E}_{Q(z_n)}\big[ \log P(z_n|\Omega_z) \big] +\mathbb{E}_{Q(W)}\big[ \log P(W|\Omega_w) \big]  \nonumber \\
 &- \sum_{n=1}^{N}\mathbb{E}_{Q(z_n)}[\log Q(z_n)]  - \mathbb{E}_{Q(W)}[\log Q(W)]
 \end{align}
 It should be noted that all terms of $\mathcal{L}(\lambda_Z, \lambda_W)$ are functions of the posterior parameters $\lambda_Z, \lambda_W$.
 \subsection{Distributed Setting}
 Consider $G=(V,E)$ as an undirected connected graph with vertices $i,j \in V$ and edges $e_{ij}=(i,j) \in E$ connecting the two vertices [2]. Each i-th node is directly connected with 1-hop neighbors in $\mathcal{B}_i = \{j|e_{ij}\in E\}$ [2]. Now, assume that each i-th node has its own set of data points $X_i = \{x_{in}|n=1,...,N_i\}$, local parameters $Z_i= \{z_{in}|n=1,...,N_i\}$ and global parameter $W_i$ where $x_{in}\in \mathbb{R}^D$ is n-th data point and $N_i$ is the number of samples collected in i-th node.  
 Each i-th node infers the approximate posterior distribution over both global and local parameters locally, based on the available data in that node. Computing the posterior distribution of the local parameters ($\{Z_i\}_{i=1}^{|V|}$) is not an issue in the distributed setting due to the fact that the posterior distribution of each local latent variable $z_n$ depends solely on the corresponding observation $x_n$ and is independent of other observations $(X_{-n})$. A naive approach for computing the global parameter ($W$) posterior distribution is to impose an additional constraint on the global parameter in each node ($W_1=W_2=...=W_{|V|}$). However, in a Bayesian framework, the parameters are random variables and the notion of equality can be replaced with equivalency. This, however, leaves several options open (e.g., strict equality, equality in distribution, or almost sure equality). Here, we propose imposing equivalency in distribution, i.e., imposing equality constraints on the parameters of the posterior distribution of the global variable in each node ($\lambda_{W_{1}} = \lambda_{W_{2}} = ... = \lambda_{W_{|V|}}$). Similar to [2], for decoupling purposes, we define a set of auxiliary variables $\rho_{ij}$ , one for each edge $e_{ij}$. This now leads to the final distributed consensus MFVI formulation that it is easy to show it is equivalent to the centralized MFVI optimization problem:
 \begin{align}\label{ooo}
[\hat{\lambda}_Z, \hat{\lambda}_W] = &\underset{\lambda_{Z_i}, \lambda_{W_i}: i \in V}{\arg\min}\; - \mathbb{E}_Q\big[ \log P(X,Z,W|\Omega_z,\Omega_w) \big] + \mathbb{E}_Q[\log Q] \nonumber \\
&s.t. \;\; \lambda_{W_i} = \rho_{ij},\;\;\rho_{ij} = \lambda_{W_j},\;\; i\in V, j\in \mathcal{B}_i
 \end{align} 
 Alternating Direction Method of Multipliers (ADMM) [17] could be used to efficiently solve the above constrained optimization problem. More precisely, ADMM ulternatively updates the variables in a block coordinate fashion by solving the augmented Lagrangian (using a linear and a quadratic penalty term) of (\ref{ooo}):
\begin{align}\label{admm45}
&[\hat{\lambda}_Z, \hat{\lambda}_{W}, \hat{\rho}, \hat{\gamma}] = \underset{\rho, \gamma, \lambda_{Z_i}, \lambda_{W_i}: i \in V}{\arg\min}\; - \sum_{i=1}^{|V|}\sum_{n=1}^{N_i} \mathbb{E}_{Q(Z_i,W_i)}\big[ \log P(x_{in}|z_{in},W_i) \big]  - \sum_{i=1}^{|V|}\sum_{n=1}^{N} \mathbb{E}_{Q(z_{in})}\big[ \log P(z_{in}|\Omega_z) \big] \nonumber \\
 &- \sum_{i=1}^{|V|}\mathbb{E}_{Q(W_i)}\big[ \log P(W_i|\Omega_w) \big]   + \sum_{i=1}^{|V|}\sum_{n=1}^{N_i}\mathbb{E}_{Q(z_{in})}[\log Q(z_{in})]  + \sum_{i=1}^{|V|}\mathbb{E}_{Q(W_i)}[\log Q(W_i)]\nonumber \\
 &+\sum_{i\in V}\sum_{j \in \mathcal{B}_i}\bigg( \gamma_{ij1}^{\top}(\lambda_{W_i} - \rho_{ij}) + \gamma_{ij2}^{\top}( \rho_{ij} - \lambda_{W_j})\bigg)+\frac{\eta}{2}\sum_{i\in V}\sum_{j \in \mathcal{B}_i}\bigg( \|\lambda_{W_i} - \rho_{ij}\|_2^2 + \|\rho_{ij} - \lambda_{W_j}\|_2^2\bigg)
 \end{align}
Using conjugate exponential family for prior and likelihood distributions, each coordinate descent update in MFVI can be done in closed form. However, the penalty terms would be quadratic in the norm difference of $(\lambda_{W_i} - \rho_{ij})$, that may result in the non-analytic updates for $\{\lambda_{W_i}\}_{i=1}^{|V|}$ (it should be noted that updating $\{\lambda_{Z_i}\}_{i=1}^{|V|}$ can still be done in closed form because they do not appear in the equality constraints). 
 To solve \ref{ooo} efficiently, we propose to use Bregman ADMM (B-ADMM) [18] rather than standard ADMM. B-ADMM simply generalizes the ADMM by replacing the quadratic penalty term by different Bregman divergences in order to exploit the structure of problems (a brief review of the B-ADMM provided in the Appendix). Since global parameters are the parameters of the natural exponential family distributions, we propose to use the \textit{log partition function} $\mathcal{A}_w(.)$ of the global parameter as the bregman function. It is worth noting that the $\mathcal{A}_w(.)$ is not a stricly convex function in general, but, it is estrictly convex if the exponential family is \textbf{minimal} (an exponential family is minimal if the functions $\eta(.)$ and the statistics $\mathcal{T}(.)$ each are linearly independent. We can always achieve this by reparametrization. Hence, using minimal representation of the exponential families, it is easy to show that using this bregman function, coordinate descent steps \ref{admm2}, and \ref{admm3} of B-ADMM for solving \ref{ooo} have a closed-form solution for most of exponential families (e.g. Normal, Gamma, Beta, Bernouli, Dirichlet). Based on the proposed Bregman function, we get the following updates for BADMM:
\begin{align}\label{admm2}
&[\lambda_Z^{(t+1)}, \lambda_{W}^{(t+1)}] = \underset{\lambda_{Z_i}, \lambda_{W_i}: i \in V}{\arg\min}\; - \sum_{i=1}^{|V|}\sum_{n=1}^{N_i} \mathbb{E}_{Q(z_{in},W_i)}\big[ \log P(x_{in}|z_{in},W_i) \big]  - \sum_{i=1}^{|V|}\sum_{n=1}^{N} \mathbb{E}_{Q(z_{in})}\big[ \log P(z_{in}|\Omega_z) \big] \nonumber \\
 &- \sum_{i=1}^{|V|}\mathbb{E}_{Q(W_i)}\big[ \log P(W_i|\Omega_w) \big]   + \sum_{i=1}^{|V|}\sum_{n=1}^{N_i}\mathbb{E}_{Q(z_{in})}[\log Q(z_{in})]  + \sum_{i=1}^{|V|}\mathbb{E}_{Q(W_i)}[\log Q(W_i)]\nonumber \\
 &+\sum_{i\in V}\sum_{j \in \mathcal{B}_i}\bigg( \gamma_{ij1}^{(t)}{}^{\top}(\lambda_{W_i} - \rho_{ij}^{(t)}) + \gamma_{ij2}^{(t)}{}^{\top}( \rho_{ij}^{(t)} - \lambda_{W_j})\bigg)+\eta\sum_{i\in V}\sum_{j \in \mathcal{B}_i} B_{\mathcal{A}_w} (\lambda_{W_i}, \rho_{ij}^{(t)})
 \end{align}
 \begin{align}\label{admm3}
\rho^{(t+1)} = \underset{\rho}{\arg\min}\; \sum_{i\in V}\sum_{j \in \mathcal{B}_i}\bigg( \gamma_{ij1}^{(t)}{}^{\top}(\lambda_{W_i}^{(t+1)} - \rho_{ij}) + \gamma_{ij2}^{(t)}{}^{\top}( \rho_{ij} - \lambda_{W_j}^{(t+1)})\bigg)+\eta\sum_{i\in V}\sum_{j \in \mathcal{B}_i} B_{\mathcal{A}_w} (\rho_{ij}, \lambda_{W_i}^{(t+1)})
 \end{align}
\begin{align}\label{admm4}
\hat{\gamma}_{ijk}^{(t+1)} = \gamma_{ijk}^{(t)} + \eta (\lambda_{W_i}^{(t+1)} - \rho_{ij}^{(t+1)})\;\;\; k=1,2,\; i \in V, j \in \mathcal{B}_i
\end{align}
where $\gamma_{ijk}, i,j \in V$ with $k=1,2$ are the Lagrange multipliers. The scalar value $\eta$ is the parameter of the ADMM algorithm that should be determined in advanced (refer [17] for more information about how to tune this parameter). $B_{\mathcal{A}_w}(.,.)$ denotes the bregman divergence induced by $\mathcal{A}_w(.)$ and is defined as:
\begin{equation*}
B_{\mathcal{A}_w}(x,y) = \mathcal{A}_w(x) - \mathcal{A}_w(y) - \langle x-y, \nabla \mathcal{A}_w(y)\rangle
\end{equation*}
where $\nabla$ denotes the gradient operator and $x^{(t)}$ denotes the value of the parameter $x$ at iteration $t$. It should be noted that since Bregman divergences are not necessarily convex in the second argument, we cannot use the same $B_{\mathcal{A}_w} (\lambda_{W_i}, \rho_{ij}^{(t)})$ for the bregman penalization term in \ref{admm3}, hence, Wang and Banerjee [18] proposed use of the bregman divergence with reverese of the parameters $(B_{\mathcal{A}_w} (\rho_{ij}, \lambda_{W_i}^{(t+1)}))$ and they proved the convergence of the new update equations.

The intuition behind the proposed bregman function is as follows: based on the fact that the bregman divergence (using log partition function as bregman function) between two parameters $\lambda$, $\lambda'$ of the same (minimal) exponential family $\mathcal{P}(x)$ is equivalent to the reverse KL divergence between the exponential families (Eq. \ref{ddd}) [23] and assuming that $\rho_{ij}$ is the natural parameter of the same exponential family as $Q_{\lambda_{W_i}}(.)$, penalizing the deviation of the posterior parameter $\lambda_{W_i}$ from the parameter $\rho_{ij}$ using the bregman divergence $B_{\mathcal{A}_w}(\lambda_{W_i}, \rho_{ij})$ is equivalent to penalizing the deviation of the approximate posterior distribution $Q_{\lambda_{W_i}}(W_i)$ from the distribution $Q_{\rho_{ij}}(.)$ in KL sense.
\begin{equation}\label{ddd}
B_{\mathcal{A}_w}(\lambda,\lambda') = \mathcal{A}_w(\lambda) - \mathcal{A}_w(\lambda') - \langle \lambda-\lambda', \nabla \mathcal{A}_w(\lambda')\rangle = KL(\mathcal{P}_{\lambda'}(x), \mathcal{P}_{\lambda}(x))
\end{equation}
\section{Case Study: Distributed Bayesian PCA (D-BPCA)}\label{BPCA}
In what follows, we derive D-MFVI in the context of Bayesian PCA [8]. Consider a data set $X$ of observed D-dimensional vectors $X = \{x_n \in \mathbb{R}^D\}_{n=1}^{N}$ with the corresponding latent variables $Z=\{z_n \in \mathbb{R}^{M}\}_{n=1}^{N}$ whose prior distribution is a zero mean Gaussian $P(z_n) = \mathcal{N}(z_n;0,I)$:
In Probabilistic PCA model, the observed variable $x_n$ is then defined as a linear transformation of $z_n$ with additive
Gaussian noise $\epsilon$: $x_n = Wz_n + \mu + \epsilon$, where $W$ is a $D\times M$ matrix, $\mu$ is a d-dimensional vector
and $\epsilon$ is a zero-mean Gaussian-distributed vector with precision matrix $\tau^{-1} I$. So, the likelihood distribution is:
\begin{equation}\label{li}
P(x_n|z_n,W,\mu,\tau) = \mathcal{N}(x_n;Wz_n+\mu,\tau^{-1}I),\;\;\; n=1,...,N
\end{equation} 
Based on the above probabilistic formulation of PCA, we can obtain a Bayesian treatment of PCA by first introducing a prior distribution $P(\mu, W, \sigma^2)$ over the parameters of the model . Second, we compute the corresponding posterior distribution $P(\mu, W, \tau|X)$ by multiplying the prior by the likelihood function given by (\ref{li}), and normalizing. 

There are two issues that must be addressed in this framework: (i) the choice of prior distribution, and (ii) the formulation of a tractable
procedure for computing the posterior distribution. Typically, in BPCA, the prior distributions over parameters are defined such that they are independent of each other apriori $P(\mu, W, \tau) = P(\mu)P(W)p(\tau)$.
For simplicity, in this paper we assume that the data noise precision $\tau$ is a fixed but unknown parameter. We define an independent Gaussian prior over each row of $W$ as
\begin{equation}\label{zxc2}
P(W| \alpha) = \prod_{d=1}^{D}(\frac{\alpha_d}{2\pi})^{M/2}exp\{-\frac{1}{2}\alpha_d (w_d - \bar{w}_d)^{\top}(w_d - \bar{w}_d) \}
\end{equation} 
where $w_d$ is the d-th row of $W$, $\{\bar{w}_d\}_{d=1}^D$ and $\{\alpha_d\}_{d=1}^D$ are the mean and the precision hyper parameters respectively. Furthermore, we consider another Gaussian distribution as prior for $\mu$: $P(\mu) = \mathcal{N}(\bar{\mu},\theta^{-1}I)$, where $\bar{\mu}$ and $\theta$ are the mean and the precision hyper parameters respectively. Due to the intractability of computing the exact posterior distribution of the parameters, we use MFVI to approximate the posterior distribution. In order to apply MFVB to Bayesian PCA we assume a fully factorized $Q$ distribution of the form
\begin{equation}
Q(W,Z,\mu) = \prod_{d=1}^{D}\prod_{m=1}^{M}Q(w_{dm})\prod_{n=1}^{N}Q(z_n)\prod_{d=1}^{D}Q(\mu_d)
\end{equation}
Due to the use of conjugate priors for $W,Z$ and $\mu$, the posterior distributions are Gaussian $(Q(w_{dm})\sim \mathcal{N}(m_{dm}^w, \lambda_{dm}^w)$, $Q(\mu_d) \sim \mathcal{N}(m_d^{\mu},\lambda_{dm}^{\mu})$, and $Q(z_n) \sim \mathcal{N}(m_n^z,\Lambda_n^{-1})$ are Gaussian and their update is equivalent to re-estimation of the corresponding means and variances.
\subsection{Distributed formulation}
The distributed MFVI algorithm developed in Section \ref{pm} can be directly applied to this BPCA model. Based on the terminology of section \ref{pm}, $W$ and $\mu$ are global latent variables, and $\{z_n\}_{n=1}^{N}$ are the local latent variables. The basic idea is to assign each subset of samples to each node in the network, and do inference locally in each node. By considering $\Xi_i = \{(m_{dm}^w)_i, (\lambda_{dm}^w)_i,(m_d^{\mu})_i,(\lambda_{dm}^{\mu})_i, (m_n^z)_i,(\Lambda_n)_i^{-1}\}$ as the set of parameters for node $i$, The D-MFVI optimization now becomes:
\begin{align}\label{obj}
\hat{\Xi} = &\underset{\Xi_i: i \in V}{\arg\min}\; - \sum_{i=1}^{|V|}\mathbb{E}_{Q(W_i,\mu_i,Z_i)}\big[ \log P(X_i,Z_i,W_i|\tau, \alpha, \theta, \bar{\mu}, \bar{w}_1,...,\bar{w}_d) \big] + \mathbb{E}_{Q(W_i,\mu_i,Z_i)}[\log Q(W_i,\mu_i,Z_i)] \nonumber \\
&s.t. \;\; (m^{\mu}_d)_i = (\rho^{\mu}_d)_{ij},\;\; (\rho^{\mu}_d)_{ij} = (m^{\mu}_d)_j,\;\; i\in V, j\in \mathcal{B}_i \nonumber \\
& s.t.\;\; (m_{dm}^{w})_i = (\rho_{dm}^{w})_{ij},\;\; (\rho_{dm}^{w})_{ij} = (m_{dm}^{w})_j,\;\; i \in V, j\in \mathcal{B}_i \nonumber \\
& s.t.\;\; (\lambda_d^{\mu})_i = (\phi_d^{\mu})_{ij},\;\; (\phi_d^{\mu})_{ij} = (\lambda_d^{\mu})_j,\;\; i \in V, j\in \mathcal{B}_i \nonumber \\
& s.t.\;\;\; (\lambda_{dm}^{w})_i = (\phi_{dm}^{w})_{ij},\;\; (\phi_{dm}^{w})_{ij} = (\lambda_{dm}^{w})_j,\;\; i \in V, j\in \mathcal{B}_i 
 \end{align} 
where $\{ (\rho^{\mu}_d)_{ij}, (\phi^{\mu}_d)_{ij}, (\rho_{dm}^{w})_{ij}, (\phi_{dm}^{w})_{ij}\}$ are auxiliary variables (we have explained how to specify hyper-parameters $\tau, \alpha, \theta, \bar{\mu}, \bar{w}_1,...,\bar{w}_d$ and presented the coordinate descent update rules for solving the above optimization problem in the Appendix). Generalizing our distributed BPCA (D-BPCA) to deal with missing data is straightforward and follows [15]. 
\section{Experimental Results}\label{er}
In this section, we first demonstrate the general convergence properties of the D-BPCA algorithm based on synthetic data. We then apply our model to a set of SfM and MF problems. For both applications, we compared our distributed algorithm with traditional SVD, Centralized PCA (C-PCA) [2], D-PPCA [2], and Centralized BPCA (C-BPCA) [15].  
\subsection{Empirical Convergence Analysis}
We generated synthetic data (using Gaussian distribution) to show the convergence of D-BPCA in various settings. Based on the results (detailed results are available in the Appendix), D-BPCA is robust to topology of the network, the number of nodes in the network, choice of the parameter $\eta$, and both data missing-at-random (MAR) and missing-not-at-random (MNAR) [15] cases. 
\subsection{D-BPCA for Structure from Motion (SfM)}
In affine SfM, based on a set of 2D points observed from multiple cameras (or views), the goal is to estimate the corresponding 3D location of those points, hence the 3D structure of the observed object as well as its motion (or, equivalently, the motion of the cameras used to view the object). A canonical way to solve this problem is factorization [14]. More precisely, by collecting all the 2D points into a measurement matrix $X$ of size $\# points \times 2\;\cdot \;\# frames$, we can factorize it into a $\# points \times 3$ 3D structure matrix $W$ and a $3 \times 2\;\cdot\;\# frames$ motion matrix $Z$. SVD can be used to find both $W$ and $Z$ in a centralized setting. C-PPCA and D-PPCA can also estimate $W$ and $Z$ using the EM algorithm [2]. Equivalently, the estimates of $W$ and $Z$ can also be found using our D-BPCA  where $W$ is treated as the latent global structure and $Z$ is the latent local camera motion (it is worth noting that D-PPCA can only provide the uncertainty around the motion matrix $Z$, while D-BPCA obtains additional estimates of the variance of the 3D structure $W$). We now show that our D-BPCA can be used as an effective framework for distributed affine SfM. For all SfM experiments, the network was connected using the ring topology, with $\eta = 10$. We equally partitioned the frames into 5 nodes to simulate 5 cameras, the convergence was set to $10^{-3}$ relative change in objective of (\ref{ooo}). We computed maximum subspace angle between the ground truth 3D coordinates and the estimated 3D structure matrix as the measure of performance (for C-BPCA and D-BPCA, we used the posterior mean of 3D structure matrix for subspace angle calculation).
\subsubsection{Synthetic Data (Cube)}
\begin{figure}[t]
\begin{center}
\includegraphics[width=0.8\linewidth]{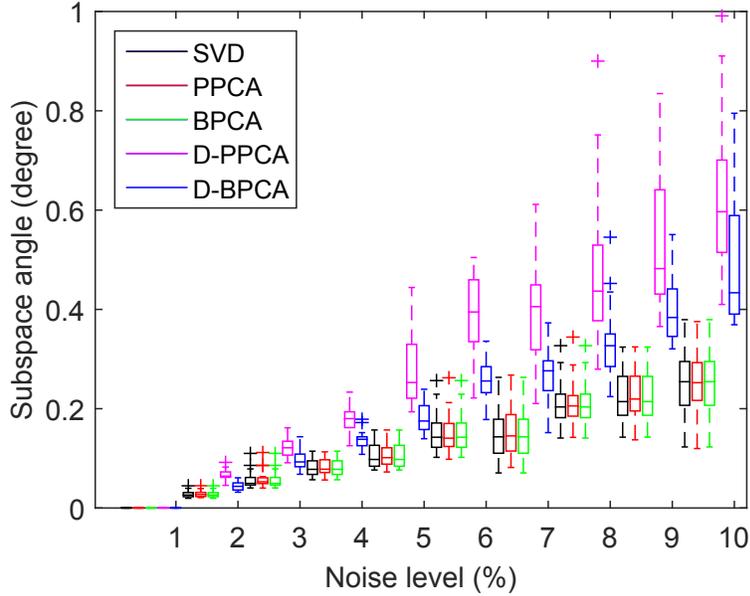}
\end{center}
\caption{Noisy data experiment for the cube synthetic data (crosses denote outliers).}
\label{fig:2a}
\end{figure}

Similar to synthetic experiments in [2], we used a rotating unit cube and 5 cameras facing the cube in a 3D space to generate synthetic data. 
In contrast to the setting in [2], we rotated the cube every $3^{\circ}$ over $150^{\circ}$ clockwise to obtain additional views necessary for our online learning evaluation i.e. in this setting, each camera observed $50$ frames. Fig. \ref{fig:2a} shows the performance of different models in the case of noisy data (over $20$ independent runs with $10$ different noise levels).

As can be seen from the figure, in the case of noisy data, D-BPCA consistently outperforms D-PPCA, thanks in part to improved robustness to overfitting.

For MAR experiment, where we randomly discarded $20\%$ of data points over ten independent runs. 
The average errors were $1.41^{\circ}$ and $1.01^{\circ}$ for D-PPCA and D-BPCA, respectively. The same experiment was done for the more challenging MNAR with the missing data generated by a realistic visual occlusion process (hence, non-random).  This yielded errors of $17.66^{\circ}$ for D-PPCA and $14.12^{\circ}$ for D-BPCA. Again, D-BPCA resulted in consistently lower errors than D-PPCA, although the error rates were higher in the more difficult MNAR setting.

\begin{figure}[t]
\begin{center}
\includegraphics[width=0.8\linewidth]{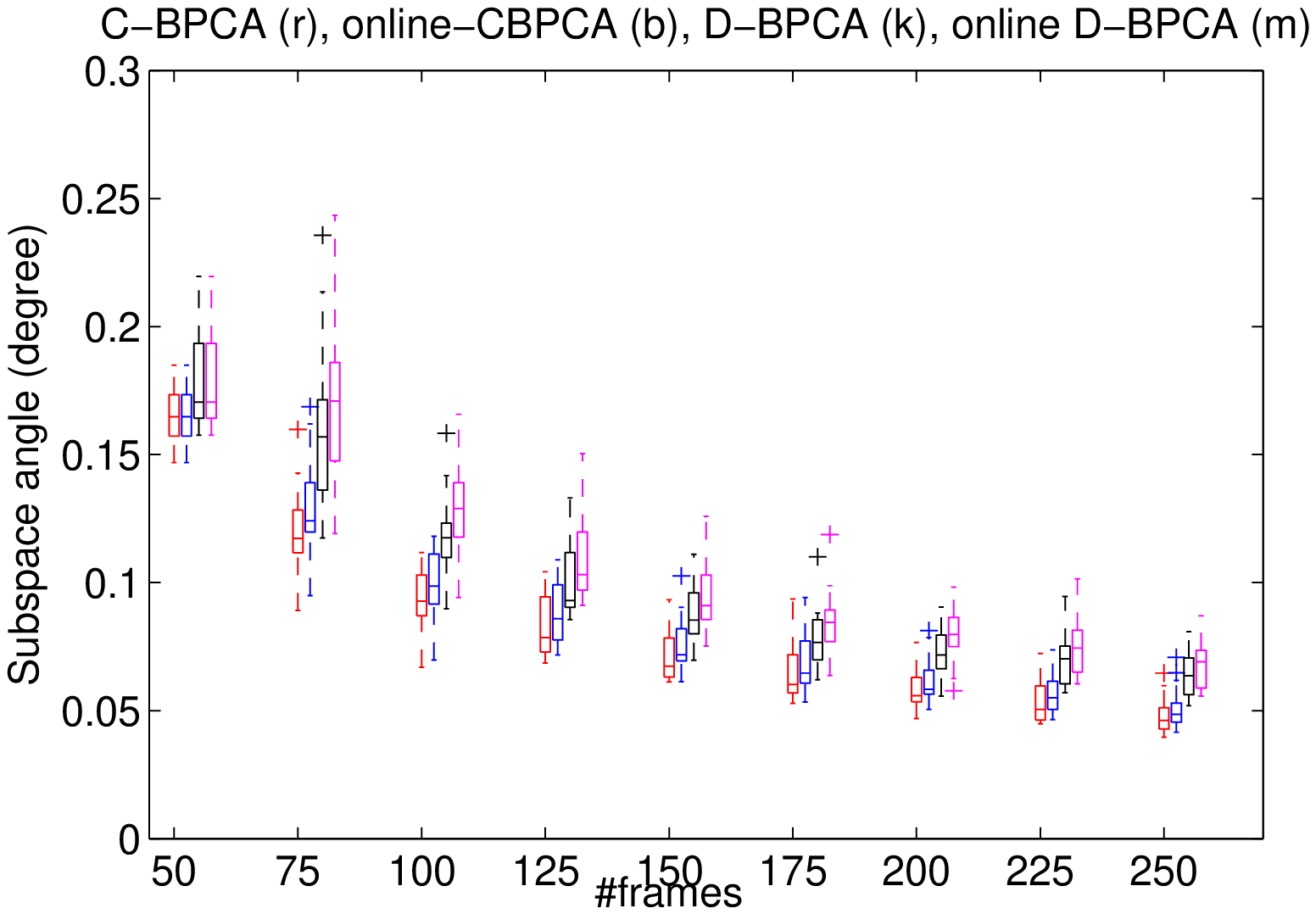}
\end{center}
\caption{Online data experiment for the cube synthetic data (crosses denote outliers).}
\label{fig:2b}
\end{figure}

One particular advantage of the D-BPCA over D-PPCA and the SVD counterparts is its ability to naturally support online Bayesian estimation in the distributed sensing network.
We first used 10 frames in each camera as the first minibatch of data. Then, we repeatedly added 5 more frames to each camera in subsequent steps. Results over 10 different runs with $1\%$ noise in the data are given in Fig.~\ref{fig:2b} (due to the non-Bayesian nature of the D-PPCA model, it cannot easily be applied in the online setting). Fig~\ref{fig:2b} demonstrates that the subspace angle error of the online D-BPCA closely follows centralized BPCA in accuracy.
\setlength{\tabcolsep}{4pt}

\begin{table}[t]
\begin{center}
\caption{Results of Caltech dataset (all results ran 20 independent initializations).}
\label{tb1}
\begin{tabular}{ c c c c c c }
\hline
\noalign{\smallskip}
Object & BallSander & BoxStuff & Rooster & Standing & StorageBin \\ 
$\#$ Points & 62 & 67 & 189 & 310 & 102 \\
$\#$ Frames & 30 & 30 & 30 & 30 & 30 \\
\hline
\noalign{\smallskip}
\multicolumn{6}{c}{Subspace angle b/w centralized SVD SfM and D-PPCA (degree)} \\
\hline
\noalign{\smallskip}
Mean & 1.4934 & 1.4402 & 1.4698 & 2.6305 & 0.4513 \\
Variance & 0.4171 & 0.4499  & 0.9511 & 1.7241 & 1.2101 \\
\hline

\noalign{\smallskip}
\multicolumn{6}{c}{Subspace angle b/w centralized SVD SfM and D-BPCA (degree)} \\
\hline
\noalign{\smallskip}
Mean & \bf 0.9910 & \bf 0.9879 & \bf 1.3855 & \bf 0.9621 & \bf 0.4203 \\
Variance & 0.0046 & 0.0986  & 0.0080 & 0.0033 & 0.0044 \\
\hline
%
\end{tabular}
\end{center}
\end{table}
\setlength{\tabcolsep}{1.4pt}

\subsubsection{Real Data}
We applied our model on the Caltech 3D Objects on Turntable dataset [19] and Hopkins155 dataset [20] to demonstrate its usefulness for real data. Following [2], we used a subset of the dataset which contains images of 5 objects for Caltech dataset, and 90 single-object sequences for Hopkins155 dataset. For both datasets, we used the same setup as [2]. The subspace angles (mean and variance of 20 independent runs) between the structure inferred using the traditional centralized SVD and the D-PPCA and D-BPCA 
 for Caltech dataset are available in Table~\ref{tb1}. As can be seen, the D-BPCA performance is better than D-PPCA (the MAR/MNAR results for the Caltech dataset and are available in the Appendix).
Average maximum subspace angle between D-PPCA, D-BPCA and SVD for all selected $90$ objects without missing data and with $10\%$ MAR are shown in Table~\ref{tb3} (we did not perform MNAR experiments on Hopkins due to the fact that the ground truth occlusion information is not provided with the dataset). Also, for this dataset, D-BPCA consistently has better performance than D-PPCA. It should be noted that although the subspace angle error is very large for MAR case for both D-PPCA and D-BPCA, 3D structure estimates were similar to that of SVD only with orthogonal ambiguity issue.
\begin{table}[t]
\begin{center}
\caption{Subspace angles (degree) between fully observable centralized SVD and D-PPCA / D-BPCA for Hopkins dataset (all results ran 5 independent initializations).}
\label{tb3}
\begin{tabular}{ l c c c c c c}
\hline
\noalign{\smallskip}
&  & No-missing & MAR \\ 
\hline
\noalign{\smallskip}
D-PPCA & Mean & 3.9523 & 13.4753  \\
& Variance & 3.3119 & 12.9832  \\
\hline
\noalign{\smallskip}
D-BPCA & Mean & \bf 0.7975 & \bf 6.4372 \\
& Variance & 0.5684 & 5.0689 \\
\hline
\noalign{\smallskip}
\end{tabular}
\end{center}
\end{table}
\setlength{\tabcolsep}{1.4pt}
\begin{table}[t]
\caption{Root mean squared error for matrix factorization experiment on Netflix dataset.}
\label{tb2}
\begin{center}
\begin{tabular}{llll}
\hline
\noalign{\smallskip}
\multicolumn{1}{c}{ C-PPCA}  &\multicolumn{1}{c}{D-PPCA} &\multicolumn{1}{c}{C-BPCA} &\multicolumn{1}{c}{D-BPCA}
\\ 
\hline
\noalign{\smallskip}
1.7723         & 1.9934  & 1.4465 & \bf 1.5613\\
\hline
\end{tabular}
\end{center}
\end{table}

\subsection{D-BPCA for Matrix Factorization (MF)}
Matrix factorization (MF) is a canonical approach for recommender systems due to its promising performance. In this section we apply our D-BPCA for large-scale MF problem using the Netflix dataset [22] that contains movie ratings given by $N = 480189$ customers to $D = 17770$ movies. The PCA model is one of the most popular techniques considered by the Netflix contestants [21]. We used given $R = 100480507$
ratings from 1 to 5, and predicted the $R' = 1408395$ ratings of the \textit{probing} set. We used 10 nodes to divide the whole rating matrix into 10 sub-blocks for local computation, where each sub-block contains $D$ rows and $N/10$ columns. We set $\eta = 10$ and the number of principle components to $12$. The root mean squared error (RMSE) of different methods (for Bayesian models, we simply used the posterior means to do reconstruction) is shown in Table~\ref{tb2}, from which we can see that by our model, local Bayesian computation effectively arrives at the solution of centralized approaches that handle all data in a single batch. In addition to lower RMSE the crucial benefit of the distributed method is that the data is distributed over different nodes, reducing storage requirements and enabling parallel computation, hence computational speedups.
\section{Discussion and Future Work}\label{co}
In this paper we introduced a general approximate inference approach using Mean Field Variational Inference for learning parameters of traditional centralized probabilistic models in a distributed setting. The main idea is to split the data into different nodes and impose the equality constraints on the posterior parameters of each node, and then, solving the constrained variational inference using Bregman ADMM. We illustrated this approach with BPCA for SfM and Matrix Factorization applications and more importantly, this algorithm generalizes to many settings. We developed our framework with MFVI and closed form coordinate updates with conjugate exponential family models. Although this class of models is pervasive, the proposed framework cannot be used in nonconjugate models. A more principled approach would be to use our framework for more advanced variational inference algorithms such as collapsed variational inference [11,12] and structured variational inference [13].

\appendix
\section*{Appendix}

\section{Bregman Alternative Direction Method of Multipliers}
ADMM has been successfully applied in a broad range of machine learning applications [17]. ADMM is canonically used for optimizing the following objective function subject to an equality constraint:
\begin{align}\label{1}
\underset{x\in \mathcal{X}, z\in \mathcal{Z}}{\arg \min} \; f(x) + g(z), \;\; s.t. \; Ax + Bz = c
\end{align}
where $f$ and $g$ are convex functions, $A,B$ and $c$ are some fixed terms. ADMM iteratively solve the the augmented Lagrangian of (\ref{1}), which is defined as follows:
\begin{align}
L_p (x,z,y) = f(x) + g(z) + \langle y, Ax + Bz -c \rangle + \eta /2 \| Ax+Bz - c \|_2^2
\end{align}
where $y$ is dual variable, $\eta >0$ is penalty parameter, and the goal of quadratic penalty term is to penalize the violation of the equality constraint. ADMM iteratively solve the above optimization problem by spliting the variables using the following updates:
\begin{align}
\label{222}
x_{t+1} &= \underset{x}{\arg\min}\; f(x) + \langle y_t, Ax + Bz_t -c \rangle + \eta /2 \| Ax+Bz_t - c \|_2^2 \\
\label{3333}
z_{t+1} &= \underset{z}{\arg\min}\; g(z) + \langle y_t, Ax_{t+1} + Bz -c \rangle + \eta /2 \| Ax_{t+1}+Bz - c \|_2^2 \\
y_{t+1} &= y_t + \eta (Ax_{t+1} + Bz_{t+1} -c)
\end{align}
Bregman ADMM (BADMM) replaces the quadratic penalty term in ADMM by a Bregman divergence [18]. More precisely, the quadratic penalty term in the x and z updates (\ref{222})-(\ref{3333}) will be replaced by a Bregman divergence in BADMM:
\begin{align}\label{444}
x_{t+1} &= \underset{x}{\arg\min}\; f(x) + \langle y_t, Ax + Bz_t -c \rangle + \eta B_{\phi}(c-Ax, Bz_t) \\
\label{555}
z_{t+1} &= \underset{z}{\arg\min}\; g(z) + \langle y_t, Ax_{t+1} + Bz -c \rangle + \eta B_{\phi}(Bz, c-Ax_{t+1}) \\
y_{t+1} &= y_t + \eta (Ax_{t+1} + Bz_{t+1} -c)
\end{align}
where $B_{\phi}: \Omega \times RI(\Omega) \rightarrow \mathbb{R}_+$ is the Bregman divergence with Bregman function $\phi$ ($\phi$ is a strictly convex function on $\Omega$) that is defined as:
\begin{align}\label{a9}
B_{\phi}(x,y) = \phi(x) - \phi(y) - \langle \nabla \phi(y), x - y\rangle 
\end{align}

\section{Full Derivation of D-MFVI for BPCA}
Before describing the proposed model and deriving the ADMM equations, we briefly describe the notations:

\begin{align*}
&(x_n)_i : \text{a} \;D\times 1 \;\text{vector which denotes the n-th data point in the i-th node} \nonumber \\
&(z_n)_i :  \text{a} \;M\times 1 \;\text{vector which denotes the latent variable of the n-th data point in the i-th node} \nonumber \\
&(\mu)_i :  \text{a} \;D\times 1 \;\text{vector which denotes the local mean in the i-th node} \nonumber \\
&(W)_i :  \text{a} \;D\times M \;\text{matrix which denotes the local weight matrix in the i-th node} \nonumber \\
&(\tau)_i :  \text{a} \;\text{scalar which denotes the local precision of the data noise in the i-th node} \nonumber \\
&(\alpha)_i :  \text{a} \;\text{scalar which denotes the local precision of the weight matrix noise in the i-th node} \nonumber \\
&(\theta)_i :  \text{a}  \;\text{scalar which denotes the local precision of the mean vector noise in the i-th node} \nonumber \\
&(m^z_n)_i :  \text{a} \;M\times 1 \;\text{vector which denotes the mean of the n-th latent variable in the i-th node} \nonumber \\
&(\Lambda^z_n)_i :  \text{a} \;M\times M \;\text{matrix which denotes the precision matrix of the n-th latent variable in the i-th node} \nonumber \\
&(m_d^{\mu})_i :  \text{a} \;scalar\;\text{which denotes the mean parameter of the d-th element of the local mean} (\mu)_i \text{in the i-th node} \nonumber \\
&(\lambda_d^{\mu})_i :  \text{a} \;scalar \;\text{which denotes the variance of the d-th element of the local mean} (\mu)_i \text{in the i-th node} \nonumber\\
&\;\;\;\;\;\;\;\;\;\;\;\text{in the i-th node(} (\lambda^{\mu}_d)_i \text{denotes the d-th element of this vector)} \nonumber \\
&(m^w_{dm})_i :  \text{a} \;scalar \;\text{which denotes the mean parameter of the (d,m)-th entry of the local weight matrix} (W)_i \text{in the i-th node} \nonumber \\
&(\lambda_{dm}^{w})_i :  \text{a} \;scalar \;\text{which denotes the variance of of the (d,m)-th entry of the local weight matrix} (W)_i \text{in the i-th node} \nonumber\\
&(\gamma_d^{\mu})_i : \text{a} \;scalar \; \text{which denotes the Lagrange multiplyer corresponding to} (m_d^{\mu})_i \nonumber \\
&(\beta_d^{\mu})_i : \text{a} \;scalar \; \text{which denotes the Lagrange multiplyer corresponding to} (\lambda^{\mu}_d)_i \nonumber \\
&(\gamma_{dm}^{w})_i : \text{a} \;scalar \; \text{which denotes the Lagrange multiplyer corresponding to} (m_{dm}^w)_i \nonumber \\
&(\beta^{w}_{dm})_i : \text{a} \;scalar \; \text{which denotes the Lagrange multiplyer corresponding to} (\lambda^{w}_{dm})_i 
\end{align*}
Prior and Liklihood distributions are as follows:
\begin{itemize}
\item $P(x_n|z_n,W,\mu,\tau) = \mathcal{N}(x_n;Wz_n+\mu,\tau^{-1}I),\;\;\;n=1,...,N$
\item $P(z_n) = \mathcal{N}(z_n;0,I),\;\;\;n=1,...,N$
\item $P(\mu) = \mathcal{N}(\mu; \bar{\mu},\theta^{-1}I)$
\item $P(W|\alpha) = \prod_{d=1}^{D}\mathcal{N}(w_{d};\bar{w}_{d},diag(\alpha^{-1}))$
\end{itemize}
where $\alpha = [\alpha_1, \alpha_2,...,\alpha_M]$.
So, the joint distribution of the parameters and the observations are as follows:
\begin{align}
P(X,Z,W,\mu |\tau, \alpha,\theta) = &\prod_{n=1}^{N}\mathcal{N}(x_n;Wz_n+\mu,\tau^{-1}I)\mathcal{N}(z_n;0,I)\prod_{d=1}^{D}\prod_{m=1}^{M}\mathcal{N}(w_{dm};\bar{w}_{dm},\alpha_{m}^{-1}) \times\nonumber \\
&\prod_{d=1}^{D}\mathcal{N}(\mu_d;\bar{\mu}_d,\theta^{-1})
\end{align}
The mean field Variational Distribution is as follows:
\begin{itemize}
\item $q(W,\mu,Z=\{z_1,...,z_N\}) = q(Z)q(W)q(\mu)$
\item $q(Z) = \prod_{n=1}^{N}\mathcal{N}(z_n;m^z_n,(\Lambda^z_n)^{-1})$
\item $q(\mu) = \prod_{d=1}^{D}\mathcal{N}(\mu;m^{\mu}_d,\lambda^{\mu}_d)$
\item $q(W) = \prod_{d=1}^{D}\prod_{m=1}^{M}\mathcal{N}(w_{dm};m^w_{dm},\lambda_{dm})$
\end{itemize}

\subsection{Notations for Distributed BPCA}
Some Notations:
\begin{itemize}
\item $G=(V,E)$: Undirected connected graph with vertices in V and edges in E
\item $i,j \in V$ : Node index
\item $e_{ij} \in E$: Edge connecting node i and node j
\item $\mathcal{B}_i=\{j|e_{ij}\in E$: Set of neighbor nodes directly connected to i-th node
\item $N_i$:The number of samples collected in i-th node
\item $(z_{n})_i$: n-th $M\times 1$ dimensional latent variable at node i where $n = 1,...,N_i$
\item $(x_{n})_i$: n-th $D\times 1$ dimensional column vector at node i where $n = 1,...,N_i$
\item $Z_i = \{(z_{n})_i;n=1,...,N_i\}$
\item $X_i = \{(x_{n})_i;n=1,...,N_i\}$
\item $\Omega_i = \{(m^z_n)_i,(\Lambda^z_n)_i,(m^{\mu})_i,(\lambda^{\mu}_d)_i, (m^w_{dm})_i, (\lambda_{dm}^w)_i\}$ : Set of local parameters of the i-th node that need to be optimized through B-ADMM algorithm.
\end{itemize}

\subsection{Distributed Algorithm for BPCA}
In this section, we derive an iterative Variational Bayes (VB) algorithm for D-BPCA using B-ADMM:
\begin{align}\label{bnb}
\hat{\Omega}_i &= \underset{\Omega_i}{argmin} \;\; -\mathbb{E}_{q_i}[\log P(X_i, Z_i, W_i,\mu_i,\tau_i,\alpha_i,\theta_i )] - H[q_i] \nonumber \\
& s.t.\;\;\; (m^{\mu}_d)_i = (\rho^{\mu}_d)_{ij},\;\; (\rho^{\mu}_d)_{ij} = (m^{\mu}_d)_j,\;\; i \in V, j\in \mathcal{B}_i \nonumber \\
& s.t.\;\;\; (\lambda_d^{\mu})_i = (\phi_d^{\mu})_{ij},\;\; (\phi_d^{\mu})_{ij} = (\lambda_d^{\mu})_j,\;\; i \in V, j\in \mathcal{B}_i \nonumber \\
& s.t.\;\;\; (m_{dm}^{w})_i = (\rho_{dm}^{w})_{ij},\;\; (\rho_{dm}^{w})_{ij} = (m_{dm}^{w})_j,\;\; i \in V, j\in \mathcal{B}_i \nonumber \\
& s.t.\;\;\; (\lambda_{dm}^{w})_i = (\phi_{dm}^{w})_{ij},\;\; (\phi_{dm}^{w})_{ij} = (\lambda_{dm}^{w})_j,\;\; i \in V, j\in \mathcal{B}_i \nonumber \\
\end{align}
where $\{ (\rho^{\mu}_d)_{ij}, (\phi^{\mu}_d)_{ij}, (\rho_{dm}^{w})_{ij}, (\phi_{dm}^{w})_{ij}, \}$ are auxiliary variables. If we solve this local optimization problem, we also solve global optimization since global optimization is simply the summation of local ones given consensus constraints meet. It should be noted that we put the same hyper-parameters for all nodes.
$\mathbb{E}_{q_i}[\log P(X_i, Z_i, W_i,\mu_i,\tau_i,\alpha_i,\theta_i)]$ can be expanded as follows:
\begin{align}
& = \frac{\tau_i}{2} \bigg(\sum_{n=1}^{N_i}(x_n)_i^T(x_n)_i -2 \sum_{n=1}^{N_i}(x_{n})_i^T \langle (W)_i\rangle \langle (z_n)_i\rangle - 2\sum_{n=1}^{N_i} (x_n)_i^T\langle (\mu)_i\rangle + 2 \sum_{n=1}^{N_i} \langle (\mu)_i\rangle ^T \langle (W)_i\rangle  \langle (z_n)_i\rangle \nonumber \\
&+ \sum_{n=1}^{N_i} \langle (\mu)_i^T (\mu)_i \rangle + \sum_{n=1}^{N_i} \langle (z_n)_i^T(W)_i^T(W)_i(z_n)_i\rangle \bigg) -\frac{N_i D}{2} \log \tau_i +\frac{1}{2}\sum_{n=1}^{N_i} \langle ((z_n)_i )^{\top}((z_n)_i ) \rangle - \frac{1}{2}\sum_{n=1}^{N_i}\log \big( |\bar{\Lambda}_n^z|\big)  \nonumber \\
&+\frac{1}{2}\sum_{d=1}^{D}\langle ((w_d)_i - (\bar{w}_d)_i)^{\top} diag(\alpha_i)((w_d)_i - (\bar{w}_d)_i)\rangle -\frac{D}{2} \sum_{m=1}^{M}\log \alpha_m +\frac{\theta_i}{2}\langle ((\mu)_i - (\bar{\mu})_i)^{\top}((\mu)_i - (\bar{\mu})_i)\rangle\nonumber \\
&     -\frac{D}{2} \log \theta_i+\frac{1}{2}\sum_{n=1}^{N_i}\log \big( |\Lambda_n^z|\big)- \sum_{d=1}^{D}\sum_{m=1}^{M}\log \big( (\lambda^w_{dm})_i \big) - \sum_{d=1}^{D}\log \big( (\lambda^{\mu}_d)_i \big)
\end{align}
where $\langle . \rangle$ denotes the expectation operator respect to the approximate distribution $q$.
$H[q_i]$ is computed as:
\begin{align}
&H[q_i] = H[q_i(Z_i)] + H[q_i(W_i)] + H[q_i(\mu_i)] \nonumber \\
& = \sum_{n=1}^{N_i}H_{q_i}[(z_n)_i] + \sum_{d=1}^{D}\sum_{j=1}^{M}H_{q_i}[(w_{dj})] + \sum_{d=1}^{D}H_{q_i}[(\mu_d)] \nonumber \\
& = -\frac{1}{2}\sum_{n=1}^{N_i}\log \big( |\Lambda_n^z|\big)+ \sum_{d=1}^{D}\sum_{m=1}^{M} \log \big( (\lambda^w_{dm})_i \big) + \frac{1}{2}\sum_{d=1}^{D}\log \big( (\lambda^{\mu}_{d})_i \big) 
\end{align}
So, the augmented Lagrangian of (\ref{bnb}) can be expressed as bellow. It should be noted that we utilized the Eq. 10 of the main paper (using KL divergence) to drive the update equations.
\begin{align*}
& = \frac{\tau_i}{2} \bigg(\sum_{n=1}^{N_i}(x_n)_i^T(x_n)_i -2 \sum_{n=1}^{N_i}(x_{n})_i^T \langle (W)_i\rangle \langle (z_n)_i\rangle - 2\sum_{n=1}^{N_i} (x_n)_i^T\langle (\mu)_i\rangle + 2 \sum_{n=1}^{N_i} \langle (\mu)_i\rangle ^T \langle (W)_i\rangle  \langle (z_n)_i\rangle \nonumber \\
&+ \sum_{n=1}^{N_i} \langle (\mu)_i^T (\mu)_i \rangle + \sum_{n=1}^{N_i} \langle (z_n)_i^T(W)_i^T(W)_i(z_n)_i\rangle \bigg) -\frac{N_i D}{2} \log \tau_i +\frac{1}{2}\sum_{n=1}^{N_i} \langle ((z_n)_i )^{\top}((z_n)_i ) \rangle - \frac{1}{2}\sum_{n=1}^{N_i}\log \big( |\bar{\Lambda}_n^z|\big)  \nonumber \\
&+\frac{1}{2}\sum_{d=1}^{D}\langle ((w_d)_i - (\bar{w}_d)_i)^{\top} diag(\alpha_i)((w_d)_i - (\bar{w}_d)_i)\rangle -\frac{D}{2} \sum_{m=1}^{M}\log \alpha_m +\frac{\theta_i}{2}\langle ((\mu)_i - (\bar{\mu})_i)^{\top}((\mu)_i - (\bar{\mu})_i)\rangle\nonumber \\
&     -\frac{D}{2} \log \theta_i+\frac{1}{2}\sum_{n=1}^{N_i}\log \big( |\Lambda_n^z|\big)- \sum_{d=1}^{D}\sum_{m=1}^{M} \log \big( (\lambda^w_{dm})_i \big) - \sum_{d=1}^{D}\log \big( (\lambda^{\mu}_d)_i \big)
\end{align*}
\begin{align}  
& +\sum_{i \in V}\sum_{j \in \mathcal{B}_i}\sum_{d=1}^{d}\bigg( (\gamma^{\mu}_d)_{ij1}\big((m^{\mu}_d)_i - (\rho^{\mu}_d)_{ij})\big)+(\gamma^{\mu}_d)_{ij2} \big((\rho^{\mu}_d)_{ij} - (m^{\mu}_d)_j \big)\bigg)\nonumber \\
&+\sum_{i \in V}\sum_{j \in \mathcal{B}_i}\bigg( \sum_{d=1}^{D}(\beta_d^{\mu})_{ij1}\big((\lambda_d^{\mu})_i - (\phi_d^{\mu})_{ij}\big)+\sum_{d=1}^{D}(\beta_d^{\mu})_{ij2}\big((\phi_d^{\mu})_{ij} - (\lambda_d^{\mu})_j\big)  \bigg)\nonumber \\
& + \sum_{i \in V}\sum_{j \in \mathcal{B}_i}\bigg( \sum_{d=1}^{D}\sum_{m=1}^{M}(\gamma_{dm}^{w})_{ij1}^{\top}\big((m_{dm}^{w})_i - (\rho_{dm}^{w})_{ij}\big)+ \sum_{d=1}^{D}\sum_{m=1}^{M}(\gamma_{dm}^{w})_{ij2}^{\top}\big((\rho_{dm}^{w})_{ij} - (m_{dm}^{w})_j\big) \bigg)\nonumber \\
& + \sum_{i \in V}\sum_{j \in \mathcal{B}_i}\bigg( \sum_{d=1}^{D}\sum_{m=1}^{M}(\beta_{dm}^{w})_{ij1}\big((\lambda_{dm}^{w})_i - (\phi_{dm}^{w})_{ij}\big)+ \sum_{d=1}^{D}\sum_{m=1}^{M}(\beta_{dm}^{w})_{ij2}\big((\phi_{dm}^{w})_{ij} - (\lambda_{dm}^{w})_j\big) \bigg)\nonumber \\
& +\eta\sum_{i \in V}\sum_{j \in \mathcal{B}_i}\bigg( \sum_{d=1}^{D}KL\big((m^{\mu}_d)_i; (\lambda^{\mu}_d)_i||  (\rho^{\mu}_d)_{ij}; (\phi^{\mu}_d)_{ij}\big) + KL\big((m^{\mu}_d)_j; (\lambda^{\mu}_d)_j||  (\rho^{\mu}_d)_{ij}; (\phi^{\mu}_d)_{ij}\big)\bigg)\nonumber \\
& +\eta\sum_{i \in V}\sum_{j \in \mathcal{B}_i}\bigg( \sum_{d=1}^{D}KL\big( (\rho^{\mu}_d)_{ij}; (\phi^{\mu}_d)_{ij} ||(m^{\mu}_d)_i; (\lambda^{\mu}_d)_i\big) + KL\big(\rho^{\mu}_{ij}; \phi^{\mu}_{ij} || (m^{\mu}_d)_j; (\lambda^{\mu}_d)_j\big)\bigg)\nonumber \\
& + \eta\sum_{i \in V}\sum_{j \in \mathcal{B}_i}\sum_{d=1}^{D}\sum_{m=1}^{M}\bigg( KL\big((m_{dm}^{w})_i;(\lambda_{dm}^w)_i ||(\rho_{dm}^{w})_{ij};(\phi_{dm}^{w})_{ij}\big) +  KL\big((m_{dm}^{w})_j;(\lambda_{dm}^w)_j ||(\rho_{dm}^{w})_{ij};(\phi_{dm}^{w})_{ij}\big)\bigg)\nonumber \\
& + \eta\sum_{i \in V}\sum_{j \in \mathcal{B}_i}\sum_{d=1}^{D}\sum_{m=1}^{M}\bigg( KL\big( (\rho_{dm}^{w})_{ij};(\phi_{dm}^{w})_{ij}|| (m_{dm}^{w})_i;(\lambda_{dm}^w)_i\big) +  KL\big( (\rho_{dm}^{w})_{ij};(\phi_{dm}^{w})_{ij}|| (m_{dm}^{w})_j;(\lambda_{dm}^w)_j\big)\bigg)\nonumber \\
\end{align}
$\{(\gamma_{dm}^{w})_{ijk}\}$, $\{(\gamma^{\mu}_d)_{ijk}\}$, $\{(\beta_{dm}^{w})_{ijk}\}$, $\{(\beta_{d}^{\mu})_{ijk}\}$,   
with $k =1,2$ are the Lagrange multipliers, and $\eta$ is a positive scalar.
We solve the above optimization problem using B-ADMM.
We cyclically minimize the above objective function $(\mathcal{L}(\Omega_i))$ over its parameters, then follow a gradient ascent step over the Lagrange multipliers. 
For deriving the update equations of the parameters, we omit $t$ temporarily for notational brevity.
\subsection*{Update for $(Z_n)_i$}
To optimise the latent parameters $(z_n)_i$, we can directly optimise Equation (\ref{bnb}). However, since
 our variational approximation are in the exponential family, and there is no ADMM constraints on these parameters, we can instead
directly give the update Equations for $(z_n)_i$  given all the rest. Hence, it is easy to show that $m_n^z$ and $\Lambda_n^z$ can be updated as (we omit $t$ notations):
\begin{equation}
(\Lambda_{n}^{z})_i = I + \tau_i \langle (W)_i^{\top}(W)_i\rangle  
\end{equation}
\begin{equation}
(m_n^z)_i = \tau_i(\Lambda_{n}^{z})_i^{-1} \langle (W)_i^{\top} \rangle ((x_n)_i - \langle(\mu)_i \rangle) 
\end{equation}
So, we have:
\begin{equation}\label{up-z1}
(\Lambda_{n}^{z})_i = I + \tau_i^{(t)} \big( 
\mathcal{M}_i^{\top}\mathcal{M}_i + \sum_{d=1}^{D}(\Lambda^w_d)_i^{-1}\big)  
\end{equation}
\begin{equation}\label{up-z2}
(m_n^z)_i = \tau_i(\Lambda_{n}^{z})_i^{-1} \mathcal{M}_i^{\top} ((x_n)_i - m^{\mu}_i) 
\end{equation}
where
\begin{align}\label{mmm}
\mathcal{M}_i &= 
		\begin{bmatrix}
	- (m^w_1)_i -\\
	\vdots \\
	- (m^w_D)_i -
	\end{bmatrix}
\end{align}

\subsection*{Update for auxilary parameters $(\rho^{\mu}_d)_{ij}$, $(\phi^{\mu}_{d})_{ij}$, $(\phi^{\mu}_d)_{ij}$, and $(\phi^{\mu}_{d})_{ij}$}
\begin{align}
(\rho^{\mu}_d)_{ij}^{(t+1)} &= \underset{(\rho^{\mu}_d)_{ij}}{argmin}\; ((\gamma^{\mu}_d)_{ij2} - (\gamma^{\mu}_d)_{ij1})(\rho^{\mu}_d)_{ij} \nonumber \\
& + \eta\bigg( KL\big((m^{\mu}_d)_i; (\lambda^{\mu}_d)_i||  (\rho^{\mu}_d)_{ij}; (\phi^{\mu}_d)_{ij} \big) + KL\big((m^{\mu}_d)_j; (\lambda^{\mu}_d)_j||  (\rho^{\mu}_d)_{ij}; (\phi^{\mu}_d)_{ij}\big)\bigg) \nonumber \\
& = \underset{\rho^{\mu}_{ij}}{argmin}\; (\gamma^{\mu}_{ij2} - \gamma^{\mu}_{ij1})\rho^{\mu}_{ij} + \eta\bigg[ (m^{\mu}_d)_i - (\rho^{\mu}_d)_{ij})^{2}/2(\phi_d^{\mu})_{ij}+ \log (\phi_d^{\mu})_{ij} - \log (\lambda_d^{\mu})_{i} + \frac{1}{2} \frac{(\lambda_d^{\mu})_{i}}{(\phi_d^{\mu})_{ij}} - \frac{1}{2}\bigg] \nonumber \\
&+ \eta\bigg[ (m^{\mu}_d)_j - (\rho^{\mu}_d)_{ij})^{2}/2(\phi_d^{\mu})_{ij}+ \log (\phi_d^{\mu})_{ij} - \log (\lambda_d^{\mu})_{j} + \frac{1}{2} \frac{(\lambda_d^{\mu})_{j}}{(\phi_d^{\mu})_{ij}} - \frac{1}{2}\bigg]
\end{align}
The above objective function is a quadratic function of $(\rho^{\mu}_d)_{ij}$ that implies a closed form update for $(\rho^{\mu}_d)_{ij}$. 
\begin{align}
(\phi^{\mu}_d)_{ij}^{(t+1)} &= \underset{(\phi^{\mu}_d)_{ij}}{argmin}\; ((\beta^{\mu}_d)_{ij2} - (\beta^{\mu}_d)_{ij1})(\phi^{\mu}_d)_{ij} \nonumber \\
&+ \eta\bigg( KL\big((m^{\mu}_d)_i; (\lambda^{\mu}_d)_i||  (\rho^{\mu}_d)_{ij}; (\phi^{\mu}_d)_{ij} \big) + KL\big((m^{\mu}_d)_j; (\lambda^{\mu}_d)_j||  (\rho^{\mu}_d)_{ij}; (\phi^{\mu}_d)_{ij}\big)\bigg) \nonumber \\
& = \underset{\phi^{\mu}_{ij}}{argmin}\; (\beta^{\mu}_{ij2} - \beta^{\mu}_{ij1})\phi^{\mu}_{ij} + \eta\bigg[ (m^{\mu}_d)_i - (\rho^{\mu}_d)_{ij})^{2}/2(\phi_d^{\mu})_{ij}+ \log (\phi_d^{\mu})_{ij} - \log (\lambda_d^{\mu})_{i} + \frac{1}{2} \frac{(\lambda_d^{\mu})_{i}}{(\phi_d^{\mu})_{ij}} - \frac{1}{2}\bigg] \nonumber \\
&+ \eta\bigg[ (m^{\mu}_d)_j - (\rho^{\mu}_d)_{ij})^{2}/2(\phi_d^{\mu})_{ij}+ \log (\phi_d^{\mu})_{ij} - \log (\lambda_d^{\mu})_{j} + \frac{1}{2} \frac{(\lambda_d^{\mu})_{j}}{(\phi_d^{\mu})_{ij}} - \frac{1}{2}\bigg]
\end{align}
We take the derivetive of the above objective function respect to $(\phi^{\mu}_d)_{ij}$ and set it to zero:
\begin{align}
&((\beta^{\mu}_d)_{ij2} - (\beta^{\mu}_d)_{ij1}) + \eta\bigg[  -\big((m^{\mu}_d)_i - (\rho^{\mu}_d)_{ij}\big)^{2}/2(\phi_d^{\mu})_{ij}^2 +\frac{1}{(\phi^{\mu}_d)_{ij}} - \frac{1}{2} \frac{(\lambda_d^{\mu})_{i}}{(\phi_d^{\mu})_{ij}^2} \bigg] \nonumber \\
+& \eta\bigg[  -\big((m^{\mu}_d)_j - (\rho^{\mu}_d)_{ij}\big)^{2}/2(\phi_d^{\mu})_{ij}^2 +\frac{1}{(\phi^{\mu}_d)_{ij}} - \frac{1}{2} \frac{(\lambda_d^{\mu})_{j}}{(\phi_d^{\mu})_{ij}^2} \bigg]=0
\end{align}
By simplifing the above equation, we have:
\begin{align}
&((\beta^{\mu}_d)_{ij2} - (\beta^{\mu}_d)_{ij1})(\phi_d^{\mu})_{ij}^2 + \eta\bigg[  -\big((m^{\mu}_d)_i - (\rho^{\mu}_d)_{ij}\big)^{2}/2 +(\phi^{\mu}_d)_{ij} - \frac{1}{2} (\lambda_d^{\mu})_{i} \bigg] =0 \nonumber \\
& + \eta\bigg[  -\big((m^{\mu}_d)_j - (\rho^{\mu}_d)_{ij}\big)^{2}/2 +(\phi^{\mu}_d)_{ij} - \frac{1}{2} (\lambda_d^{\mu})_{j} \bigg] 
\end{align}
This is a quadratic function of $(\phi^{\mu}_d)_{ij}$ for which we can find an algebraic solution. It is easy to show that the form of update equations for $(\rho^{w}_{dm})_{ij}$ and $(\phi^{w}_{dm})_{ij}$ are the same as $(\rho^{\mu}_{d})_{ij}$ and $(\phi^{\mu}_{d})_{ij}$ respectively. Hence, we can find an algebraic solution for them.

\subsection*{Update for $(m^{\mu}_d)_i$}
\begin{align}
(m^{\mu}_d)_i^{(t+1)} &= \underset{(m^{\mu}_d)_i}{argmin}\; \frac{\tau_i}{2}\bigg[ - 2(x_{dn})_i (m^{\mu}_d)_i + 2\sum_{n=1}^{N_i}(m^{\mu}_d)_i(m^{w}_d)_i^{\top}(m_n^z)_i + N_i(m^{\mu}_d)_i^2 + \frac{\theta_i}{2}((m^{\mu}_d)_i - (\bar{m}^{\mu}_d)_i)^2
\bigg]\nonumber \\
&  +2\sum_{j \in \mathcal{B}_i}\bigg( (\gamma^{\mu}_d)_{ij1}\big((m^{\mu}_d)_i - (\rho^{\mu}_d)_{ij})\big) \big)\bigg) \nonumber \\
&+2\eta \sum_{j \in \mathcal{B}_i}\bigg[ ((m^{\mu}_d)_i - (\rho^{\mu}_d)_{ij})^{2}/2(\lambda_d^{\mu})_{j}+ \log (\lambda_d^{\mu})_{j} - \log (\phi_d^{\mu})_{ij} + \frac{1}{2} \frac{(\phi_d^{\mu})_{i}}{(\lambda_d^{\mu})_{i}} - \frac{1}{2}\bigg] \nonumber \\
\end{align}
Again, this is a quadratic function of $(m^{\mu}_d)_i$ for which we can find an algebraic solution. 

\subsection*{Update for $(m^{w}_{dm})_i$}
\begin{align}
(m^{w}_d)_i^{(t+1)} &= \underset{(m^{w}_d)_i}{argmin}\; \frac{\tau_i}{2}\bigg[ - 2(x_{dn})_i (m^{z}_{nm})_i + 2\sum_{n=1}^{N_i}(m^{\mu}_d)_i(m^{z}_{nm})_i +\sum_{m'=1}^{M}(z_{m'n})_i(z_{mn})_i(m_{dm})_i(m_{dm'})_i
\bigg]\nonumber \\
& + \frac{\alpha_m}{2}( (m^{w}_{dm})_i - (\bar{m}^{w}_{dm})_i )^2 + \sum_{j \in \mathcal{B}_i}\bigg( (\gamma_{dm}^{w})_{ij1}^T\big((m_{dm}^{w})_i - (\rho_{dm}^{w})_{ij}\big)+ (\gamma_{dm}^{w})_{ij2}^T\big((\rho_{dm}^{w})_{ij} - (m_{dm}^{w})_j\big) \bigg)\nonumber \\
&+ \eta\bigg[ (m^{w}_{dm})_i - (\rho^{w}_{dm})_{ij})^{2}/2(\phi_{dm}^{w})_{ij}+ \log (\phi_{dm}^{w})_{ij} - \log (\lambda_{dm}^{w})_{i} + \frac{1}{2} \frac{(\lambda_{dm}^{w})_{i}}{(\phi_{dm}^{w})_{ij}} - \frac{1}{2}\bigg] \nonumber \\
&+ \eta\bigg[ (m^{w}_{dm})_j - (\rho^{w}_{dm})_{ij})^{2}/2(\phi_{dm}^{w})_{ij}+ \log (\phi_{dm}^{w})_{ij} - \log (\lambda_{dm}^{w})_{j} + \frac{1}{2} \frac{(\lambda_{dm}^{w})_{j}}{(\phi_{dm}^{w})_{ij}} - \frac{1}{2}\bigg]
\end{align}
Again, this is a quadratic function of $(m^{\mu}_d)_i$ for which we can find an algebraic solution. 
\subsection*{Update for $(\lambda^{\mu}_{d})_i$}
\begin{align}
(\lambda^{\mu}_d)_i^{(t+1)} &= \underset{(\lambda^{\mu}_d)_i}{\arg\min}\;\frac{\tau_i}{2}N_i(\lambda_d^{\mu})_i + \frac{\theta_i}{2}(\lambda_d^{\mu})_i - \log (\lambda_d^{\mu})_i +\sum_{j \in \mathcal{B}_i}\bigg( (\beta_d^{\mu})_{ij1}\big((\lambda_d^{\mu})_i - (\phi_d^{\mu})_{ij}\big)+(\beta_d^{\mu})_{ij2}\big((\phi_d^{\mu})_{ij} - (\lambda_d^{\mu})_j\big)  \bigg)\nonumber \\
&+2\eta \sum_{j \in \mathcal{B}_i}\bigg[ ((m^{\mu}_d)_i - (\rho^{\mu}_d)_{ij})^{2}/2(\lambda_d^{\mu})_{ij}+ \log (\lambda_d^{\mu})_{i} - \log (\phi_d^{\mu})_{ij} + \frac{1}{2} \frac{(\phi_d^{\mu})_{i}}{(\lambda_d^{\mu})_{i}} - \frac{1}{2}\bigg] \nonumber \\
\end{align}
We take the derivetive of the above objective function respect to $(\lambda^{\mu}_d)_i$ and set it to zero:
\begin{align}
&\frac{\tau_i}{2}N_i+ \frac{\theta_i}{2} - \frac{1}{(\lambda_d^{\mu})_i} +\sum_{j \in \mathcal{B}_i}\bigg( (\beta_d^{\mu})_{ij1}-(\beta_d^{\mu})_{ij2} \bigg)\nonumber \\
&+2\eta \sum_{j \in \mathcal{B}_i}\bigg[ -((m^{\mu}_d)_i - (\rho^{\mu}_d)_{ij})^{2}/2(\lambda_d^{\mu})_{i}^2+ \frac{1}{(\lambda_d^{\mu})_{i}} - \frac{1}{2} \frac{(\phi_d^{\mu})_{i}}{(\lambda_d^{\mu})_{i}^2} \bigg] = 0 \nonumber \\
\end{align}
By simplifing the above equation, we have:
\begin{align}
&\frac{\tau_i}{2}N_i(\lambda_d^{\mu})_i^2+ \frac{\theta_i}{2}(\lambda_d^{\mu})_i^2 - (\lambda_d^{\mu})_i +\sum_{j \in \mathcal{B}_i}\bigg( (\beta_d^{\mu})_{ij1}-(\beta_d^{\mu})_{ij2} \bigg)(\lambda_d^{\mu})_i^2\nonumber \\
&+2\eta \sum_{j \in \mathcal{B}_i}\bigg[ -((m^{\mu}_d)_i - (\rho^{\mu}_d)_{ij})^{2}+ (\lambda_d^{\mu})_{i} - \frac{1}{2} (\phi_d^{\mu})_{i} \bigg] \nonumber \\
\end{align}
 This is a quadratic function of $(\lambda_d^{\mu})_i$ for which we can find an algebraic solution. 

\subsection*{Update for $(\lambda^{w}_{dm})_i$}
\begin{align}
(\lambda^{w}_{dm})_i^{(t+1)} &= \underset{(\lambda^{w}_{dm})_i}{\arg\min}\;\frac{\tau_i}{2}\sum_{n=1}^{N_i}(m_{nm}^z)_i(\lambda_{dm}^{w})_i + \frac{(\alpha_m)_i}{2}(\lambda_{dm}^{w})_i  - \log (\lambda_{dm}^{w})_i \nonumber \\
&+\sum_{j \in \mathcal{B}_i}\bigg( (\beta_{dm}^{w})_{ij1}\big((\lambda_{dm}^{w})_i - (\phi_{dm}^{w})_{ij}\big)  \bigg)\nonumber \\
&+2\eta \sum_{j \in \mathcal{B}_i}\bigg[ ((m^{w}_{dm})_i - (\rho^{w}_{dm})_{ij})^{2}/2(\lambda_{dm}^{w})_{i}+ \log (\lambda_{dm}^{w})_{i} - \log (\phi_{dm}^{w})_{ij} + \frac{1}{2} \frac{(\phi_{dm}^{w})_{i}}{(\lambda_{dm}^{w})_{i}} - \frac{1}{2}\bigg] \nonumber \\
\end{align}
We take the derivetive of the above objective function respect to $(\lambda^{w}_{dm})_i$ and set it to zero:
\begin{align}
 &\frac{\tau_i}{2}\sum_{n=1}^{N_i}(m_{nm}^z)_i + \frac{(\alpha_m)_i}{2}  - \frac{1}{(\lambda_d^{\mu})_i} +\sum_{j \in \mathcal{B}_i}\bigg( (\beta_{dm}^{w})_{ij1}  \bigg)\nonumber \\
&+2\eta \sum_{j \in \mathcal{B}_i}\bigg[ -((m^{w}_{dm})_i - (\rho^{w}_{dm})_{ij})^{2}/2(\lambda_{dm}^{w})_{i}^2+  \frac{1}{(\lambda_{dm}^{w})_{i}} - \frac{1}{2} \frac{(\phi_{dm}^{w})_{i}}{(\lambda_{dm}^{w})_{i}^2} \bigg]  = 0\nonumber \\
\end{align}
By simplifing the above equation, we have:
\begin{align}
 &\frac{\tau_i}{2}\sum_{n=1}^{N_i}(m_{nm}^z)_i(\lambda_d^{\mu})_i^2 + \frac{(\alpha_m)_i}{2}(\lambda_d^{\mu})_i^2  - (\lambda_d^{\mu})_i +\sum_{j \in \mathcal{B}_i}\bigg( (\beta_{dm}^{w})_{ij1}(\lambda_d^{\mu})_i^2  \bigg)\nonumber \\
&+2\eta \sum_{j \in \mathcal{B}_i}\bigg[ -((m^{w}_{dm})_i - (\rho^{w}_{dm})_{ij})^{2}+
(\lambda_{dm}^{w})_{i} - \frac{1}{2}(\phi_{dm}^{w})_{i} \bigg]  = 0\nonumber \\
\end{align}
This is a quadratic function of $(\lambda^{w}_{dm})_i$ for which we can find an algebraic solution. 
\begin{equation}
(\gamma^{\mu}_d)_{ij1}^{(t+1)} = (\gamma^{\mu}_d)_{ij1}^{(t)} + \eta \big((m^{\mu}_d)_i^{(t+1)} - (\rho^{\mu}_d)_{ij}^{(t+1)}\big),\;\forall i \in V, j \in \mathcal{B}_i
\end{equation}
\begin{equation}
(\gamma^{\mu}_d)_{ij2}^{(t+1)} = (\gamma^{\mu}_d)_{ij2}^{(t)} + \eta \big((\rho^{\mu}_d)_{ij}^{(t+1)} - (m^{\mu}_d)_j^{(t+1)} \big) \;\forall i \in V, j \in \mathcal{B}_i
\end{equation}
\begin{equation}
(\beta_d^{\mu})_{ij1}^{(t+1)} = (\beta_d^{\mu})_{ij1}^{(t)} + \eta\big((\lambda_d^{\mu})_i^{(t+1)} - (\phi_d^{\mu})_{ij}^{(t+1)}\big), \;\forall i \in V, j \in \mathcal{B}_i
\end{equation}
\begin{equation}
(\beta_d^{\mu})_{ij2}^{(t+1)} = (\beta_d^{\mu})_{ij2}^{(t)} + \eta\big((\phi_d^{\mu})_{ij}^{(t+1)} - (\lambda_d^{\mu})_j^{(t+1)}\big), \;\forall i \in V, j \in \mathcal{B}_i
\end{equation}
\begin{equation}
(\gamma_{dm}^{w})_{ij1}^{(t+1)} = (\gamma_{dm}^{m})_{ij1}^{(t)} + \eta \big((m_{dm}^{w})_i^{(t+1)} - (\rho_{dm}^{w})_{ij}^{(t+1)}\big),\;\forall i \in V, j \in \mathcal{B}_i
\end{equation}
\begin{equation}
(\gamma_{dm}^{w})_{ij2}^{(t+1)} = (\gamma_{dm}^{w})_{ij2}^{(t)} + \eta \big((\rho_{dm}^{w})_{ij}^{(t+1)} - (m_{dm}^{w})_j^{(t+1)} \big),\;\forall i \in V, j \in \mathcal{B}_i
\end{equation}
\begin{equation}
(\beta_{dm}^{w})_{ij1}^{(t+1)} = (\beta_{dm}^{w})_{ij1}^{(t)} + \eta\big((\lambda_{dm}^{w})_i^{(t+1)} - (\phi_{dm}^{w})_{ij}^{(t+1)}\big), \;\forall i \in V, j \in \mathcal{B}_i
\end{equation}
\begin{equation}
(\beta_d^{\mu})_{ij2}^{(t+1)} = (\beta_d^{\mu})_{ij2}^{(t)} + \eta\big((\phi_d^{\mu})_{ij}^{(t+1)} - (\lambda_d^{\mu})_j^{(t+1)}\big) ,\;\forall i \in V, j \in \mathcal{B}_i
\end{equation}

\section{Learning the hyper-parameters}
Instead of pre-specifying the local hyper-parameters $\tau_i$, $\alpha_i$, and $\theta_i$, we put local Gamma prior distribution on them $(\theta_i \sim \Gamma( a_{\theta}, b_{\theta}), \alpha_{im}\sim \Gamma( a_{\alpha}, b_{\alpha}), \tau_i \sim \Gamma( a_{\tau},b_{\tau}))$ and try to learn them through our coordinate decsent parameter updates (we use the Gamma distribution as the approximate posterior distribution of the hyper-parameters: $q(\alpha_{im}) = \Gamma( a'_{\alpha_{im}}, b'_{\alpha_{im}}), q(\theta) = \Gamma(a'_{\theta_i},b'_{\theta_i}), q(\tau_i) = \Gamma(a'_{\tau_i},b'_{\tau_i})$). Since there are no B-ADMM constraints on these parameters, their closed-form mean field updates can be easily obtained. However, it should be noted that for updating each of the above update equations, in which the hyper-parameters appear, the hyper-paramerters should be replaced by their approximate posterior means. In other words, $\tau_i$, $\alpha_{im}$, and $\theta_i$ should be replaced by $\frac{a'_{\tau_i}}{2b'_{\tau_i}}$, $\frac{a'_{\alpha_{im}}}{2b'_{\alpha_{im}}}$, and $\frac{a'_{\theta_i}}{2b'_{\theta_i}}$ respectively.

\section{Synthetic Data Result}
\begin{figure}[t]
\begin{center}
\begin{subfigure}[b]{0.3\textwidth}
       \includegraphics[width=\textwidth]{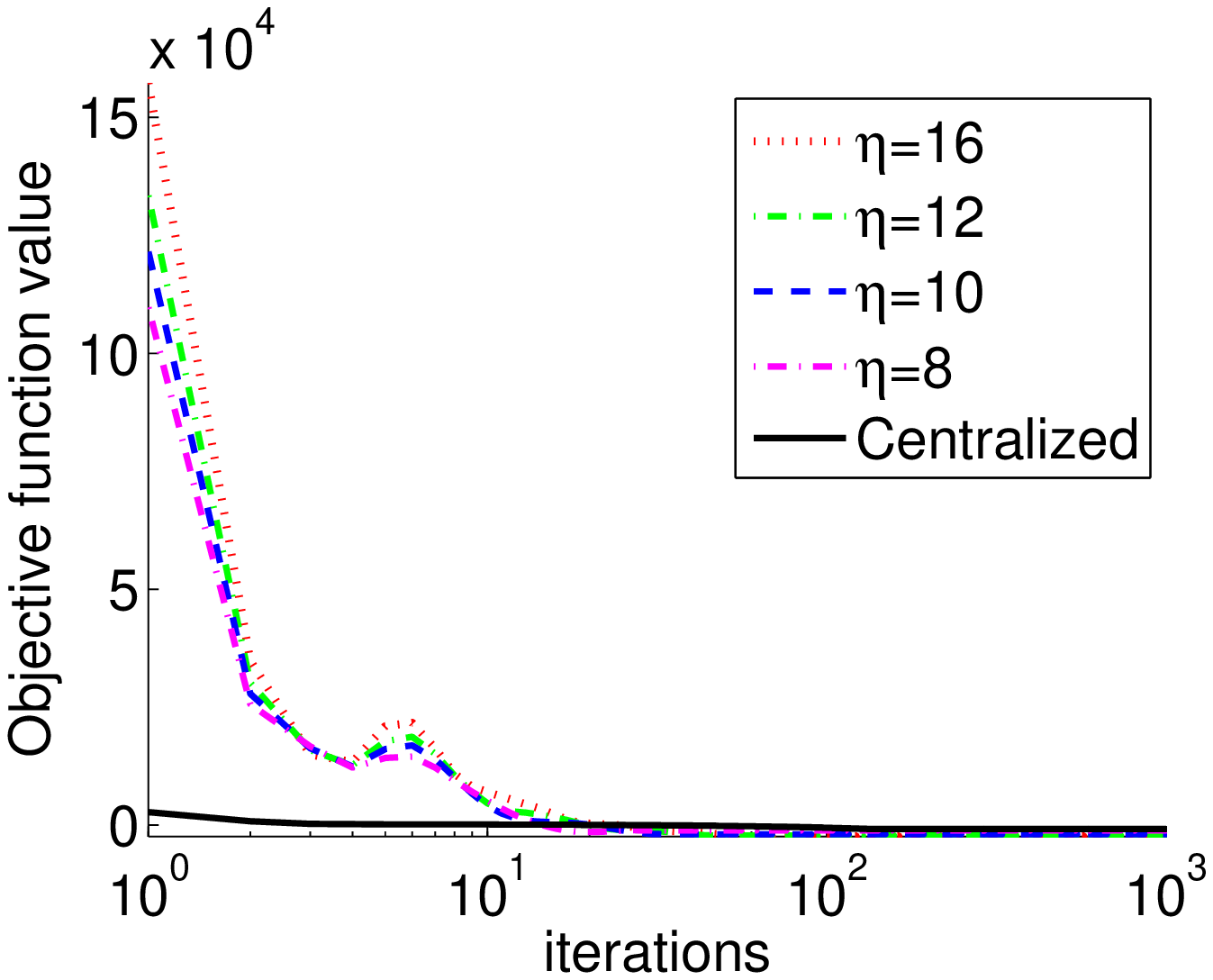}
       \caption{Impact of $\eta$}
       \label{fig:11}
   \end{subfigure}
   ~ 
   \begin{subfigure}[b]{0.3\textwidth}
       \includegraphics[width=\textwidth]{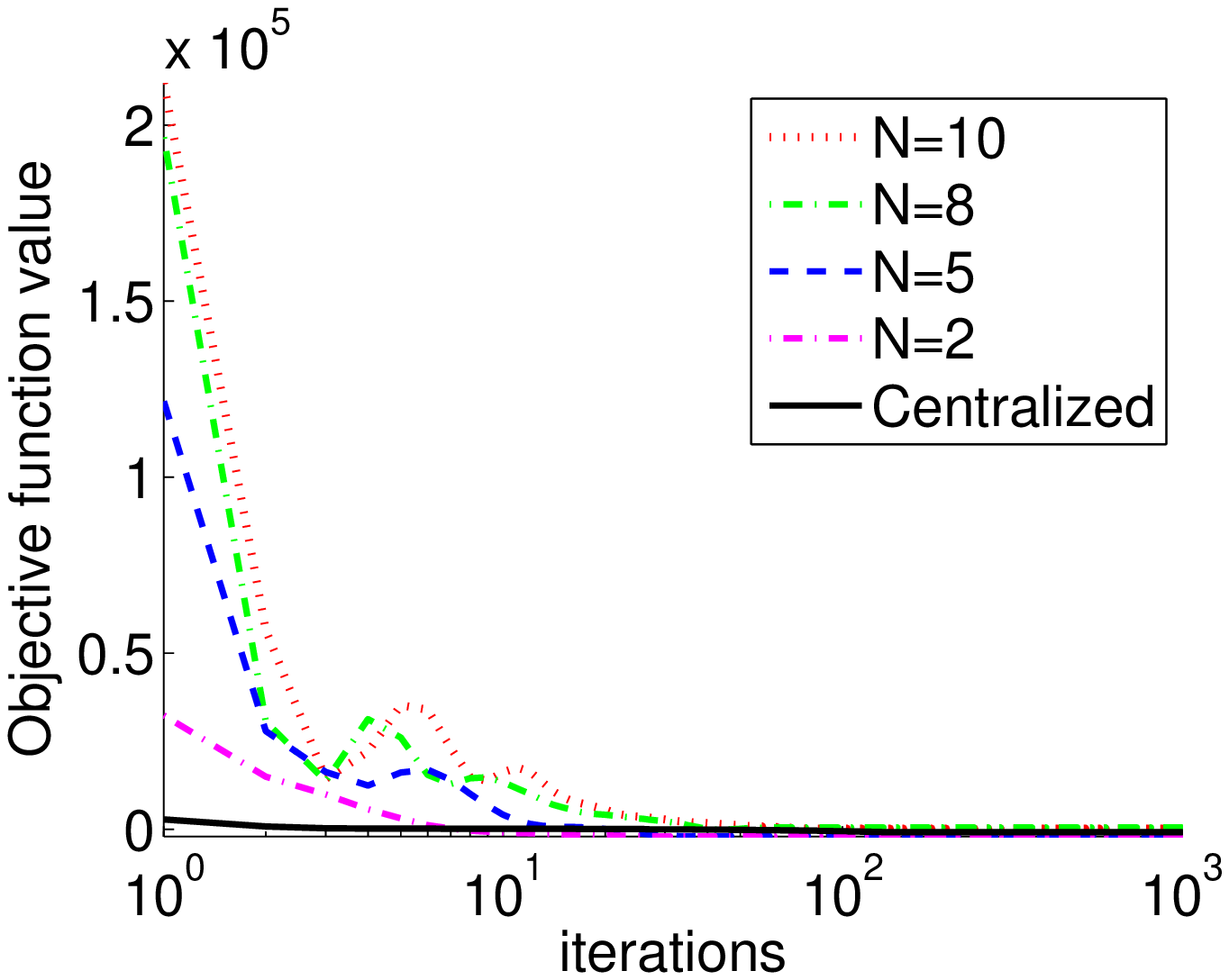}
       \caption{Impact of number of nodes}
       \label{fig:2}
   \end{subfigure}
   ~ 
   \begin{subfigure}[b]{0.3\textwidth}
       \includegraphics[width=\textwidth]{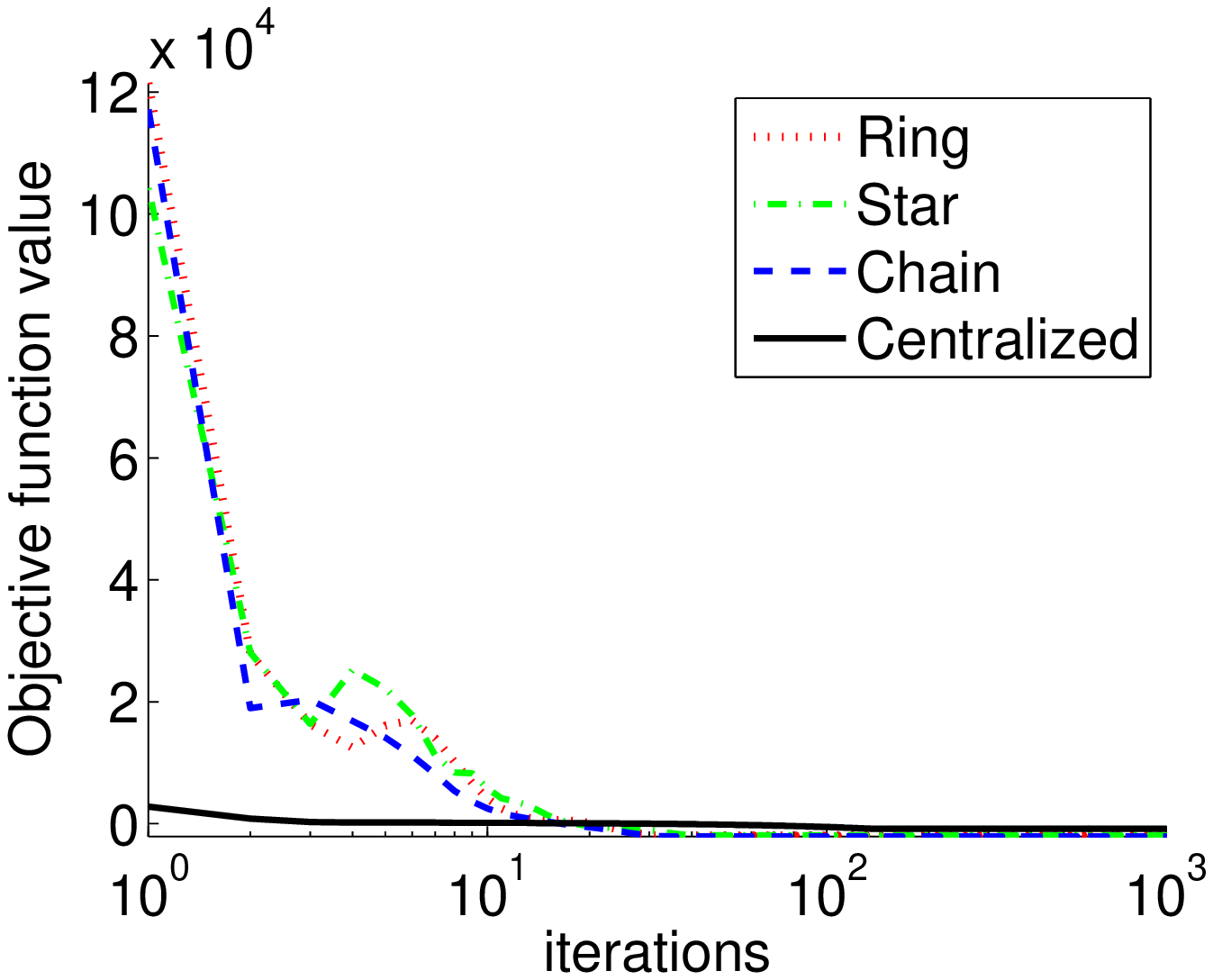}
       \caption{Impact of topologies}
       \label{fig:3}
\end{subfigure}
\end{center}
\end{figure}

In this section, we show the empirical convergence properties of the D-BPCA. We generated 50 dimensional 250 random samples from $\mathcal{N}(5,0.8)$. We assigned 50 samples equally to each node in a 5-nodes network connected with ring topology to find a 5 dimensional subspace. Our convergence criterion is the relative change in objective of (6) in the main paper and we stop when it is smaller than
$10^{-3}$. We initialized parameters with random values from a uniform distribution. All the results are averaged over 20
independent random initializations. \\
Fig.~\ref{fig:11} shows the convergence curve of D-BPCA for various $\eta$ values. As can be seen, all $\eta$ values lead to convergence within $10^2$ iterations. Furthermore, the value they converge to is equivalent to centralized solution meaning we can achive the same global solution using the distributed algorithm. Fig.~\ref{fig:2} shows convergence curve as a function of the number of nodes in a network. In all cases, D-BPCA successfully converged within $10^2$ iterations. Similar trends were observed with networks of more than 10 nodes. We also conducted experiments to test the effects of network topology on the parameter convergence. Fig.~\ref{fig:3} depicts the result for three simple network types. In all cases we considered, D-BPCA reached near the stationary point within only $10$ iterations. 

%
%

\begin{figure}[t]
\begin{center}
\includegraphics[width=0.7\linewidth]{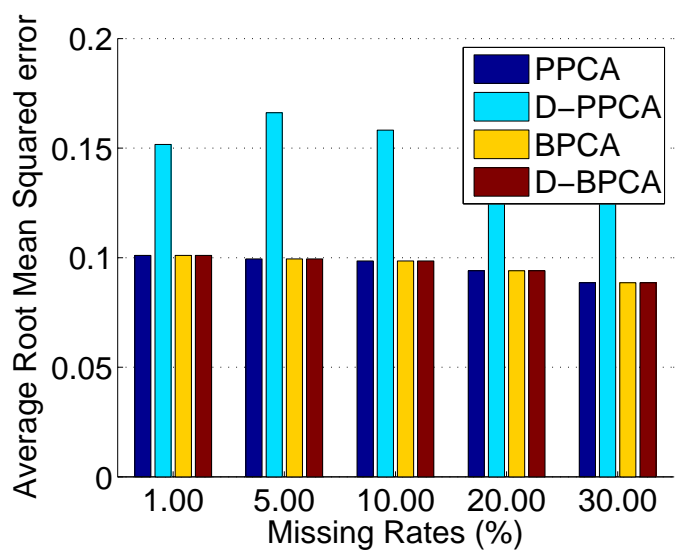}
\end{center}
\caption{Missing At Random experiment: Average root mean squared error of reconstructions based on PPCA, BPCA, D-PPCA, D-BPCA results.}
\label{fig:4-a}
\end{figure}

We consider two possibilities of missing values; the case when values missing at random (MAR) and the case when values missing not at random (MNAR). In [15], it has been shown that Bayesian formulations of PCA can deal with missing values more efficiently than probabilistic (non-Bayesian) PCA, particularly in the MAR setting. The same conclusion holds for D-BPCA. As shown in Fig.~\ref{fig:4-a}, D-BPCA has less reconstruction error than D-PPCA and can effectively reconstruct the original measurement comparable to its centralized counterpart under different amounts of missing values. This fact also holds for MNAR case although the error tends to be slightly larger than in the MAR case.
\begin{figure}[t]
\begin{center}
\includegraphics[width=0.7\linewidth]{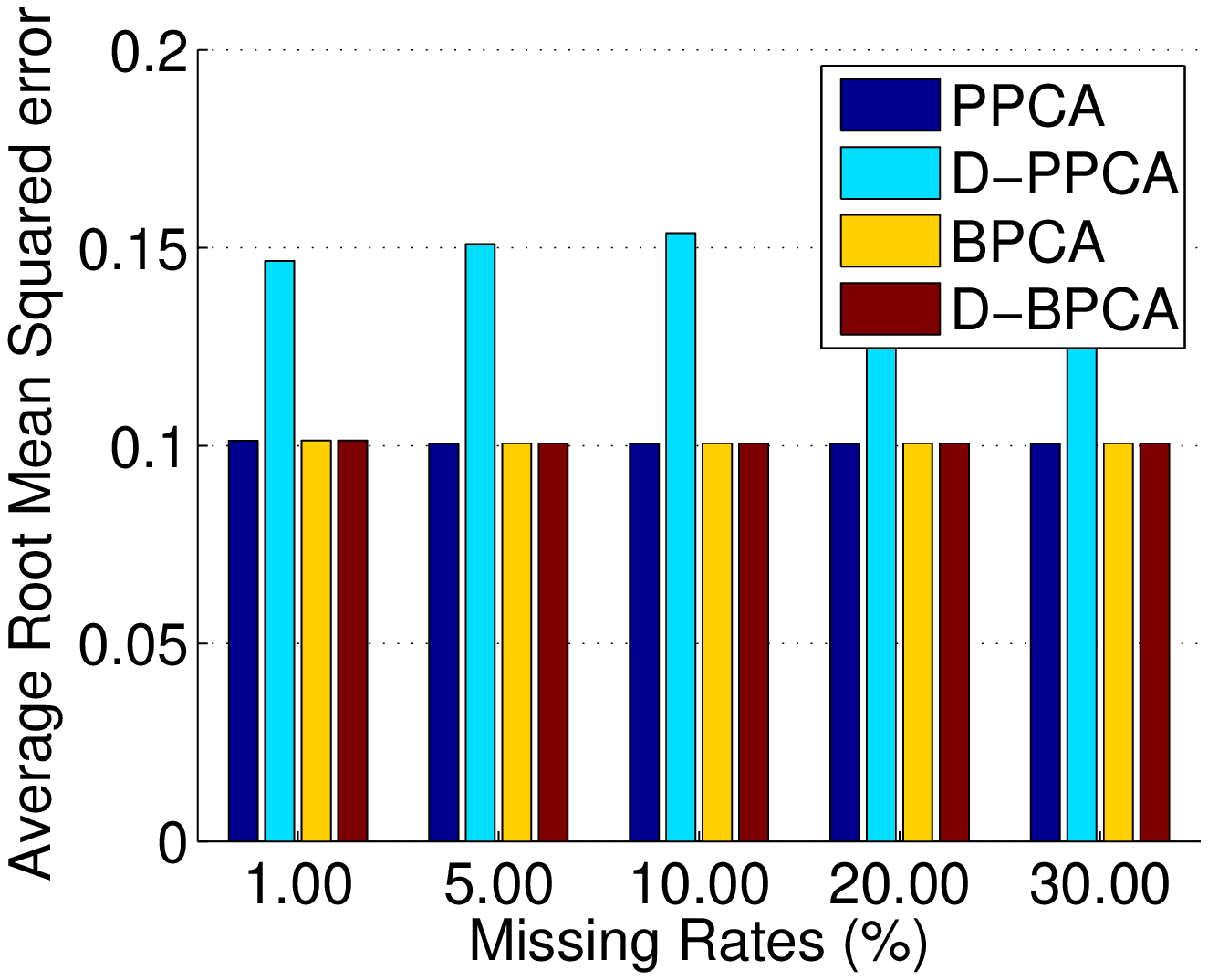}
\end{center}
\caption{Missing Not At Random experiment: Average root mean squared error of reconstructions based on PPCA, BPCA, D-PPCA, D-BPCA results.}
\label{fig:4-b}
\end{figure}

\section{Further results for Caltech Dataset}
$10\%$ MAR and MNAR results (averaged over all 5 objects) are provided in the Table~\ref{tb11}. Again, for both MAR and MNAR cases, the D-BPCA has better performance than D-PPCA. 
\begin{table}[t]
\begin{center}
\caption{Missing data Results of Caltech dataset (all results ran 20 independent initializations). Results provide variances over various missing value settings. Numbers are subspace angles between fully observable centralized SVD result versus D-PPCA / D-BPCA result.
}
\label{tb11}
\begin{tabular}{ l c c c c c c}
\hline
\noalign{\smallskip}
&  & MAR & MNAR \\ 
\hline
\noalign{\smallskip}
D-PPCA & Mean & 4.0609 & 9.4920  \\
& Variance & 1.2976 & 5.9624  \\
\hline
\noalign{\smallskip}
D-BPCA & Mean & \bf 2.2012 & \bf7.2187 \\
& Variance & 1.3179 & 5.2853 \\
\hline
\noalign{\smallskip}
\end{tabular}
\end{center}
\end{table}
\setlength{\tabcolsep}{1.4pt}

\clearpage

\section*{References}
$[1]$ Radke, R.J., (2008) A survey of distributed computer vision algorithms. {\it Handbook of Ambient Intelligence and Smart Environments}, Springer-Verlag.\\
$[2]$ Yoon, S., \& Pavlovic, V. (2012) Distributed probabilistic learning for camera networks with missing data. In F. Pereira, C. Burges,
L. Bottou, and K. Weinberger (eds.), {\it Advances in Neural Information Processing Systems 25}, pp. 2924-2932.\\
$[3]$ Broderick, T. \& Boyd, N. \& Wibisono, A. \& Wilson, A. \& Jordan, M.I. (2013) Streaming Variational Bayes. {\it  Advances in Neural Information Processing Systems 26}, pp. 1727-1735.\\
$[4]$ Giannakis, G. \& Ling, q. \& Mateos, G. \& Schizas, I. \& Zhu, H. (2015) Decentralized learning for wireless communications and networking. {\it arXiv:1503.08855v1 [math.OC]}. \\
$[5]$ Tron, R. \& Vidal, R. (2011) Distributed computer vision algorithms. {\it IEEE Signal Processing Magazine}.\\
$[6]$ Blei, D. \& Ng, A. \& Jordan, M. (2003) Latent Dirichlet allocation. {\it Journal of Machine Learning Research}, pp. 993-1022.\\
$[7]$ Attias, H. (2000) A variational Bayesian framework for graphical models. {\it In Neural Information Processing Systems}.\\
$[8]$ Ghahramani, Z. \& Beal, M. (2000) Variational inference for Bayesian mixtures of factor analysers. {\it In
Neural Information Processing Systems}. \\
$[9]$ Fox, E. \& Sudderth, E. \& Jordan, M. \& Willsky, A. (2011) A sticky HDP-HMM with application to speaker
diarization. {\it Annals of Applied Statistics}, Vol:5, 1020-1056.\\
$[10]$ Paisley, J. \& Carin, L. (2009) Nonparametric factor analysis with beta process priors. {\it In International Conference on Machine Learning}.\\
$[11]$ Teh, Y. \& Newman, D. \& Welling, M. (2006) A collapsed variational Bayesian inference algorithm for
latent Dirichlet allocation. {\it Neural Information Processing Systems}. \\
$[12]$ Teh, Y. \& Kurihara, K. \& Welling, M. (2007) Collapsed variational inference for HDP. {\it Neural Information
Processing Systems}.\\
$[13]$ Gershman, S. \& Hoffman, M. \& Blei, D. (2012) Nonparametric variational inference. {\it International
Conference on Machine Learning}. \\
$[14]$  Tomasi, C. \& Kanade, T. (1992) Shape and motion from image streams under orthography: a factorization
method. {\it International Journal of Computer Vision}, pp. 137-154.\\
$[15]$ Ilin, A. \& Raiko, T. (2010) Practical Approaches to Principal Component Analysis in the Presence
of Missing Values. {\it Journal of Machine Learning Research}, pp. 1957-2000. \\
$[16]$ Hoffman, M. \$ Blei, D. \& Wang, C. \& Paisly, J. (2013) Stochastic Variational Inference. {\it Journal of Machine Learning Research}, Vol:14, pp. 1303-1347.\\
$[17]$ Boyd, S. \& Parikh, N. \& Chu, E. \& Peleato, P. \& Eckstein, J. (2011) Distributed Optimization
and Statistical Learning via the Alternating Direction Method of Multipliers. {\it Foundations and Trends in Machine Learning}, Vol: 3, pp. 1-122. \\
$[18]$ Wang, H. \& Banerjee, A. (2014) Bregman Alternating Direction Method of Multipliers. {\it Advances in Neural Information Processing Systems 27 (NIPS 2014)}.\\
$[19]$ Moreels, P. \& Perona, P. (2007) Evaluation of Features Detectors and Descriptors based on 3D Objects.
{\it International Journal of Computer Vision}, Vol:73, pp. 263-284.\\
$[20]$ Tron, R., Vidal, R. (2007) A Benchmark for the Comparison of 3-D Motion Segmentation Algorithms.
{\it In IEEE International Conference on Computer Vision and Pattern Recognition}.\\
$[21]$ Salakhutdinov, R. \& Mnih, A. (2008) Probabilistic matrix factorization.{\it In Advances in Neural Information
Processing Systems 20}.\\
$[22]$ Bell, R. \& Koren, Y. \& Volinsky, C. (2007) The BellKor solution to the Netflix Prize. {\it Available at http: //www.netflixprize.com/}.\\
$[23]$ Davis, J. \& Kulis, B. \& Sra, S. \& Dhillon, I. (2006) Information-Theoretic Metric Learning . {\it Neural Information Processing Systems
Workshop on Learning to Compare Examples}.\\
$[24]$ Zilberstein, S. \& Russell, S. (1993) Anytime sensing, planning and action: A practical model for robot control. {\it IJCAI}, pp. 1402-1407.\\

\end{document}